%% file: main.tex
\documentclass{article}
\input{math_commands}

    \PassOptionsToPackage{numbers, compress}{natbib}
    \usepackage[final]{neurips_2024}

\usepackage[utf8]{inputenc} % allow utf-8 input
\usepackage[T1]{fontenc}    % use 8-bit T1 fonts
\usepackage{hyperref}       % hyperlinks
\hypersetup{
    colorlinks=true,
    linkcolor=blue,
    citecolor=blue,
    filecolor=magenta,      
    urlcolor=cyan,
}
\usepackage{url}            % simple URL typesetting
\usepackage{booktabs}       % professional-quality tables
\usepackage{amsfonts}       % blackboard math symbols
\usepackage{nicefrac}       % compact symbols for 1/2, etc.
\usepackage{microtype}      % microtypography
\usepackage{xcolor}         % colors
\usepackage{graphicx}

\usepackage{amssymb,amsmath,amsthm,enumitem}
\usepackage{mathtools}
\theoremstyle{plain}
\newtheorem{theorem}{Theorem}[section]
\newtheorem{proposition}[theorem]{Proposition}

\theoremstyle{definition}
\newtheorem{definition}[theorem]{Definition}

\theoremstyle{remark}

\usepackage{algorithm}
\usepackage{algorithmic}
\usepackage{titlesec}
\titlespacing*{\subsection}{0pt}{0\baselineskip}{0\baselineskip}
\titlespacing*{\subsubsection}{0pt}{0\baselineskip}{0\baselineskip}
\titlespacing*{\section}{5pt}{0\baselineskip}{0\baselineskip}
\everydisplay{
\setlength{\abovecaptionskip}{2pt}
\setlength{\abovedisplayskip}{3pt}
\setlength{\belowdisplayskip}{2pt}
\setlength{\belowcaptionskip}{0pt}
}
\usepackage{caption}
\usepackage{subcaption}
\usepackage{multirow}
\title{\method: Permutation-invariant Autoregressive Diffusion \\for Graph Generation}

\author{%
  Lingxiao Zhao\\
  Carnegie Mellon University\\
  \texttt{lingxiaozlx@gmail.com}\\
  \And
  Xueying Ding\\
  Carnegie Mellon University\\
  \texttt{xding2@andrew.cmu.edu}\\
  \And
  Leman Akoglu\\
  Carnegie Mellon University\\
  \texttt{lakoglu@andrew.cmu.edu} \\ 
}

\begin{document}

\maketitle

\begin{abstract}
\input{0-abstract}

\end{abstract}

\section{Introduction}

\input{1-introduction}

\section{Related Work}
\input{2-related}

\section{Permutation-invariant Autoregressive Denoising Diffusion}
\label{sec:pard}
\input{3-Pard}

\section{Architecture Improvement}
\label{sec:realization}

\input{4-realization}

\section{Experiments}
\label{sec:experiment}
\input{5-experiments}

\section{Conclusion}
\label{sec:conclusion}
\input{6-conclusion}

\clearpage
\bibliographystyle{plainnat}
\bibliography{reference}

\clearpage
\appendix
\section{Appendix}

\input{8-appendix}

\newpage
\section*{NeurIPS Paper Checklist}

\input{9-checklist}

\end{document}

%% file: math_commands.tex
\usepackage{amsmath,amsfonts,bm,xspace}

\newcommand{\method}{{\sc Pard}\xspace}

\newcommand{\qm}{{\sc QM9}\xspace}
\newcommand{\zinc}{{\sc ZINC250k}\xspace}
\newcommand{\moses}{{\sc MOSES}\xspace}

\newcommand{\comm}{{\sc Community-small}\xspace}
\newcommand{\cave}{{\sc Caveman}\xspace}
\newcommand{\cora}{{\sc Cora}\xspace}
\newcommand{\breast}{{\sc Breast}\xspace}

\newcommand{\grid}{{\sc Grid}\xspace}

\newcommand{\cbit}{\begin{compactitem}}
\newcommand{\ceit}{\end{compactitem}}
\newcommand{\cben}{\begin{compactenum}}
\newcommand{\ceen}{\end{compactenum}}

\newcommand{\beq}{\begin{equation}}
	\newcommand{\eeq}{\end{equation}}
 
\def\eqref#1{Eq.~(\ref{#1})}
\def\1{\bm{1}}

\newcommand{\cL}{\mathcal{L}}

\def\rve{{\mathbf{e}}}

\def\rvm{{\mathbf{m}}}

\def\rvv{{\mathbf{v}}}

\def\rvx{{\mathbf{x}}}

\def\rmG{{\mathbf{G}}}

\def\ermE{{\textnormal{E}}}

\def\ermG{{\textnormal{G}}}
\def\ermH{{\textnormal{H}}}

\def\ermV{{\textnormal{V}}}

\def\ve{{\bm{e}}}

\def\vm{{\bm{m}}}

\def\vp{{\bm{p}}}

\def\vv{{\bm{v}}}

\def\vx{{\bm{x}}}

\def\mA{{\bm{A}}}
\def\mB{{\bm{B}}}

\def\mM{{\bm{M}}}

\def\mO{{\bm{O}}}
\def\mP{{\bm{P}}}

\def\mX{{\bm{X}}}

\DeclareMathAlphabet{\mathsfit}{\encodingdefault}{\sfdefault}{m}{sl}
\SetMathAlphabet{\mathsfit}{bold}{\encodingdefault}{\sfdefault}{bx}{n}

\def\gB{{\mathcal{B}}}

\def\gE{{\mathcal{E}}}

\def\gS{{\mathcal{S}}}

\def\gV{{\mathcal{V}}}

\def\sP{{\mathbb{P}}}

\def\sR{{\mathbb{R}}}

\newcommand{\E}{\mathbb{E}}
\newcommand{\Ls}{\mathcal{L}}

\newcommand{\KL}{D_{\mathrm{KL}}}

%% file: 0-abstract.tex
Graph generation has been dominated by autoregressive models due to their simplicity and effectiveness, despite their sensitivity to node ordering. Diffusion models, on the other hand, have garnered increasing attention as they offer comparable performance %to autoregressive methods 
while being permutation-invariant. Current graph diffusion models generate graphs in a one-shot fashion, however they require extra features and thousands of denoising steps to achieve optimal performance. We introduce \method, a \underline{P}ermutation-invariant \underline{A}uto\underline{R}egressive \underline{D}iffusion model that integrates diffusion models with autoregressive methods. \method harnesses the effectiveness and efficiency of the autoregressive model while maintaining permutation invariance without order sensitivity. Specifically, we show that contrary to sets, elements in a graph are not entirely unordered and there is a unique partial order for nodes and edges. With this partial order, \method generates a graph in a block-by-block, autoregressive fashion, where each block's probability is conditionally modeled by a shared diffusion model with an equivariant network. To ensure efficiency while being expressive, we further propose a higher-order graph transformer, which integrates transformer with  PPGN. Like GPT, we extend the higher-order graph transformer to support parallel training of all blocks. Without any extra features, \method achieves state-of-the-art performance on molecular and non-molecular datasets, and scales to large datasets like MOSES containing 1.9M molecules. \method is open-sourced  at  
{\small \url{https://github.com/LingxiaoShawn/Pard}}

%% file: 1-introduction.tex
% Graph generation overview
Graphs provide a powerful abstraction for representing relational information in many domains, 
including social networks, biological and molecular structures, recommender systems, 
and networks of various infrastructures such as computers, roads, etc. 
Accordingly, generative models of graphs that learn the underlying graph distribution from data 
find applications in network science \cite{bonifati2020graph}, drug discovery \cite{li2018multi,tong2021generative}, 
protein design \cite{anand2018generative,trinquier2021efficient}, and various use-cases for Internet of Things \cite{de2022deep}. 
Importantly, they serve as a prerequisite for building a generative foundation model \citep{foundation_model} for graphs. 

% Hardness of graph generation 

Despite significant progress in generative models for images and language, graph generation is uniquely challenged by its inherent combinatorial nature. Specifically: 1) Graphs are naturally high-dimensional and \textit{discrete} with \textit{varying sizes}, contrasting with the continuous space and fixed-size advancements that cannot be directly applied here; 2) Being permutation-invariant objects, graphs require modeling an \textit{exchangeable probability} distribution, where permutations of nodes and edges do not alter a graph's probability; and 3) The rich substructures in graphs necessitate an expressive model capable of capturing \textit{higher-order} motifs and interactions. Several graph generative models have been proposed to address (part of) these challenges, based on various techniques like 
autoregression \cite{GraphRNN, GRAN}, VAEs \citep{GraphVAE}, GANs \citep{MolGAN}, flow-based methods \citep{GraphAF}, and denoising diffusion \cite{Diffussion, DiGress}. Among these, autoregressive models and diffusion models stand out with superior performance, thus significant popularity. However, current autoregressive models, while efficient,  are sensitive to node/edge order with non-exchangeable probabilities; whereas diffusion models, though promising, 
are less efficient, requiring thousands of denoising steps and extra node/edge/graph-level features (structural and/or domain-specific) to achieve high generation quality.  

% Problem of autoregressive method and diffusion method.
In this paper, we introduce \method (leopard in Ancient Greek), the \textit{first} \underline{P}ermutation-invariant \underline{A}uto\underline{R}egressive \underline{D}iffusion model that combines the efficiency of autoregressive methods and the quality of diffusion models together, while retaining the property of exchangeable probability. Instead of generating an entire graph directly, we explore the direction of generating through \textit{block-wise} graph enlargement. 
Graph enlargement offers a fine-grained control over graph generation, which can be particularly advantageous for real-world applications that require local revisions to generate graphs. 
Moreover, it essentially decomposes the joint distribution of the graph into a series of simpler conditional distributions, thereby leveraging the data efficiency characteristic of autoregressive modeling. We also argue that graphs, unlike sets, inherently exhibit a \textit{unique partial order} among nodes, naturally facilitating the decomposition of the joint distribution. Thanks to this unique partial order, \method's block-wise autoregressive sequence is permutation-invariant, unlike any prior graph autoregressive methods in the literature.

To model the conditional distribution of nodes and edges within a block, we have, for the first time, identified a fundamental challenge in equivariant models for generation: it is impossible for \textit{any} equivariant model, no matter how powerful, to perform general graph transformations without symmetry breaking. However, through a diffusion process that injects noise, a permutation equivariant network can progressively denoise to realize targeted graph transformations. This approach is inspired by the annealing process where energy is initially heightened before achieving a stable state, akin to the process of tempering iron.
Our analytical findings naturally lead to the design of our proposed \method that combines autoregressive approach with local block-wise discrete denoising diffusion. Using a diffusion model with equivariant networks ensures that each block's conditional distribution is exchangeable. Coupled with the permutation-invariant block sequence, this renders the entire process permutation-invariant and the joint distribution exchangeable. What is more, this inevitable combination of autoregression and diffusion successfully combines the strength of both approaches while getting rid of their shortcomings: By being permutation-invariant, a key advantage of diffusion, it generalizes better than autoregressive models while also being much more data-efficient. By decomposing the challenging joint probability into simpler conditional distributions, an advantage of autoregression, it requires significantly fewer diffusion steps, outperforming pure diffusion methods by a large margin. Additionally, each inference step in the diffusion process incurs lower computational cost by processing only the generated part of the graph, rather than the entire graph. And it can further leverage caching mechanisms (to be explored in future)  to avoid redundant computations.

Within \method, we further propose several architectural improvements. First, to achieve 2-FWL expressivity with improved memory efficiency, we propose a higher-order graph transformer that integrates transformer with PPGN \cite{PPGN}, while utilizing a significantly reduced representation size for edges. Second, to ensure training efficiency without substantial overhead compared to the original diffusion model, we design a GPT-like causal mechanism to support parallel training of all blocks. These extensions are generalizable and can lay the groundwork for a higher-order GPT.

\method achieves new SOTA performance on many molecular and non-molecular datasets \textit{without any extra features}, significantly outperforming DiGress \cite{DiGress}. Thanks to efficient architecture and parallel training, \method scales to large datasets like MOSES \cite{moses} with 1.9M graphs. Finally, not only \method can serve as a generative foundation model for graphs in the future, its autoregressive parallel mechanism can further be combined with language models for language-graph generative pretraining. 
% planting seeds for high-potential future work.

%{\small{\url{https://anonymous.4open.science/r/AutoregressiveGraphDiffusion-C816/}}}.

%% file: 2-related.tex
\textbf{Autoregressive (AR) Models for Graph Generation.}
AR models create graphs step-by-step, adding nodes and edges sequentially. This method acknowledges graphs' discrete nature but faces a key challenge as there is no inherent order in graph generation. To address this, various strategies have been proposed to simplify orderings and approximate the marginalization over permutations; i.e. $p(G) = \sum_{\pi \in \mathcal{P}(G)} p(G, \pi)$. \citet{li2018learning} propose using random or deterministic empirical orderings. GraphRNN \citep{GraphRNN} aligns permutations with breadth-first-search (BFS) ordering, with a many-to-one mapping. GRAN \citep{GRAN} offers marginalization over a family of canonical node orderings, including node degree descending, DFS/BFS tree rooted at the largest degree node, and k-core ordering. {GraphGEN \cite{goyal2020graphgen} uses a single canonical node ordering, but does not guarantee the same canonical ordering during generation.} \citet{order_matters} {avoid defining ad-hoc orderings by}  modeling the conditional probability of orderings,  
$p(\pi|G)$,  with a trainable AR model, estimating marginalized probabilities during training to enhance both the generative model and the ordering probability model.

\textbf{Diffusion Models for Graph Generation.} 
EDP-GNN \cite{Diffussion} is the first work that adapts score matching \cite{song2019generative} to graph generation, by viewing graphs as matrices with continuous values.  GDSS \cite{GDSS} generalizes EDP-GNN by adapting SDE-based diffusion \cite{sde-song} and considers node and edge features. 
\citet{SwinGNN} argues that learning exchangeable probability with equivariant networks is hard, hence proposes permutation-sensitive SwinGNN with continuous-state score matching. Previous works apply continuous-state diffusion to graph generation, ignoring the natural discreteness of graphs.
DiGress \cite{DiGress} is the first to apply discrete-state diffusion \cite{D3PM,hoogeboom2021argmax} to graph generation and achieves significant improvement. However, DiGress relies on many additional structural and domain-specific features. GraphArm \cite{GraphARM} applies Autoregressive Diffusion Model (ADM) \cite{AutoregressiveDiffusion} to graph generation, where exactly one node and its adjacent edges decay to the absorbing
states at each forward step based on a \textit{random} node order. Similar to AR models, GraphArm is permutation sensitive. 

We remark that although both are termed ``autoregressive diffusion'', it is important to distinguish that \method is \textit{not} ADM. The term ``autoregressive diffusion'' in our context refers to the integration of autoregressive methods with diffusion models. In contrast, ADM represents a specific type of discrete denoising diffusion  where exactly
one dimension decays to an absorbing state at a time in the forward diffusion process. 
See \citet{graphdiffusion-survey} for a survey of recent diffusion models on graphs. 
% - Diffusion models on graph

% - Autoregressive models on graph
% - 

%% file: 3-pard.tex
We first introduce setting and notations. We focus on graphs with categorical features.  
Let $G = (\gV, \gE)$ be a \textit{labeled} graph with the number of distinct node and edge labels denoted  $K_v$ and $K_e$, respectively. Let $\vv^i \in \{0,1\}^{K_v}, \forall i \in \gV$  be the one-hot encoding  of node $i$'s label. 
Let $\ve^{i,j} \in \{0,1\}^{K_e}, \forall i,j \in \gV$  
be the one-hot encoding  of the label for the edge between node $i$ and $j$. We also represent ``absence of edge'' as a type of edge label, hence $|\gE| =|\gV|\times |\gV| $. 
Let $\ermV \in \{0,1\}^{|\gV| \times K_v}$ and $\ermE \in \{0,1\}^{|\gV| \times |\gV| \times K_e }$ be the collection of one-hot encodings of all nodes and edges using the \textit{default node order}, and let $\ermG:=(\ermV, \ermE)$. To describe probability, let $\rvx$ be a random variable with its sampled value $\vx$. Similarly, $\rmG$ is a random graph with its sampled graph $\ermG$.  
In diffusion process, noises are injected from $t$$=$$0$ to $t$$=$$T$ with $T$ being the maximum time step. Let $\rvx_0 \sim p_{\text{data}}(\rvx_0)$ be the random variable of observed data with underlying distribution $p_{\text{data}}(\rvx_0)$,   $\rvx_t \sim q({\rvx_t})$ be the random variable at time $t$, and  
let $\rvx_{t|s} \sim q(\rvx_t| \rvx_s)$ denote the conditional random variable. 
Also, we interchangeably use $q(\rvx_t| \rvx_s)$, $q(\rvx_t$$=$$\vx_t| \rvx_s$$=$$\vx_s)$, and $q_{t|s}(\vx_t| \vx_s)$ when there is no ambiguity.
We model the \textit{forward diffusion} process independently for each node and edge, while the \textit{backward denoising} process is modeled jointly for all nodes and edges. 
All vectors are column-wise vectors. Let $\langle \cdot, \cdot \rangle $ denote inner product. 

\subsection{Discrete Denoising Diffusion on Graphs} \label{ssec:diffusion}

Denoising Diffusion is first developed by \citet{sohl2015deep} and later improved by \citet{DDPM}. It is further generalized to discrete-state case by \citet{hoogeboom2021argmax} and \citet{D3PM}. Taking a graph $\ermG_0$ as example, diffusion model defines a forward diffusion process to gradually inject noise to all nodes and edges independently until all reach a non-informative state $\ermG_T$. Then, a denoising network is  trained to reconstruct $\ermG_0$ from the noisy sample $\ermG_t$ at each time step, by optimizing a Variational Lower Bound (VLB) for $\log p_\theta(\ermG_0)$. Specifically, the forward process is defined as a Markov chain with $q(\ermG_{t}| \ermG_{t-1}), \forall t \in [1,T]$, and the backward denoising process is parameterized with another Markov chain $p_\theta(\ermG_{t-1}| \ermG_{t}), \forall t\in [1,T]$. Note that while the forward process is independently applied to all elements, the backward process is coupled together with conditional independence assumption. Formally, 
\begin{equation}
    \scalebox{0.97}{$
\begin{aligned}
    q(\ermG_{t}| \ermG_{t-1}) = \prod_{i\in \gV} q(\rvv^i_t | \rvv^i_{t-1}) \prod_{i,j \in \gV} q(\rve^{i,j}_t | \rve^{i,j}_{t-1}),  \quad
    p_\theta(\ermG_{t-1}| \ermG_{t}) = \prod_{i \in \gV} p_\theta(\rvv^i_{t-1} | \ermG_{t}) \prod_{i,j \in \gV} p_\theta(\rve^{i,j}_{t-1} | \ermG_{t})  \;. \label{eq:q} 
\end{aligned}    
    $}
\end{equation}
Then, the VLB  of $\log p_\theta(\ermG_0)$ can be written (see Apdx.\S\ref{ssec:vlbderive}) as
\begin{equation}
\scalebox{0.8}{$
\begin{aligned}
     &\log p_{\theta}(\ermG_0) 
     % =\log \int q(\ermG_{1:T}|\ermG_0) \frac{p_{\theta}(\ermG_{0:T})}{q(\ermG_{1:T}|\ermG_0)}d\ermG_{1:T} \\
     \geq \underbrace{\E_{q(\ermG_{1}|\ermG_0)}\big[ \log p_{\theta}(\ermG_{0}| \ermG_{1} )\big]}_{-\Ls_1(\theta)} 
      - \underbrace{\KL\big(q(\ermG_{T} |\ermG_0 ) || p_{\theta}(\ermG_{T}) \big)}_{\Ls_{\text{prior} }}
  -\sum_{t=2}^T \underbrace{\E_{q(\ermG_{t}|\ermG_0)}\big[ \KL\big(q(\ermG_{t-1} | \ermG_{t}, \ermG_0 ) || p_{\theta}(\ermG_{t-1}| \ermG_{t} \big)  \big]}_{\Ls_t(\theta)} \; 
\end{aligned}
$}\label{eq:vlb} 
\end{equation}
where  $\Ls_{\text{prior} }\approx 0$, since $p_{\theta}(\ermG_{T}) \approx q(\ermG_{T} |\ermG_0 )$ is designed as a fixed noise distribution  that is easy to sample from. To compute \eqref{eq:vlb}, we need to formalize the distributions (\textcolor{orange}{$i$}) 
 $q(\ermG_t| \ermG_0)$ and (\textcolor{orange}{$ii$})  $q(\ermG_{t-1} | \ermG_t, \ermG_0)$, as well as (\textcolor{orange}{$iii$}) the parameterization of $p_{\theta}(\ermG_{t-1}| \ermG_{t} )$. 
 DiGress \cite{DiGress} applies D3PM \cite{D3PM} to define these three terms. Different from DiGress, we closely follow the approach in \citet{zhao2024improving} to define these three terms, as their formulation is simplified with improved memory usage and loss computation. For brevity, we refer readers to Appx. \S\ref{appdx:discrete-diffusion} for the details. Notice that while a neural network can directly be  used to parameterize (\textcolor{orange}{$iii$}) $p_{\theta}(\ermG_{t-1}| \ermG_{t} )$,  we follow \cite{zhao2024improving} to parameterize $p_\theta( \ermG_0| \ermG_t)$ instead, and compute (\textcolor{orange}{$iii$}) from $p_\theta( \ermG_0| \ermG_t)$.

With (\textcolor{orange}{$i \sim iii$}) known, one can compute the negative VLB loss in \eqref{eq:vlb} exactly. In addition, at each time step $t$, the cross entropy (CE) loss  between (\textcolor{orange}{$i$}) $q(\ermG_t | \ermG_0)$ and $p_\theta(\ermG_0|\ermG_t)$ that quantifies  reconstruction quality is often employed as an auxiliary loss, which is formulated as

\vspace{-0.1in}
\begin{minipage}{\textwidth}
\begin{align}
  \Ls_t^{CE}(\theta) = -\E_{q(\ermG_t | \ermG_0)} \Big[ \sum_{i\in \gV} \log p_\theta(\rvv^i_0 | \ermG_t ) + \sum_{i,j \in \gV} \log p_\theta(\rve_0^{i,j} | \ermG_t )
  \Big] \;.   \nonumber \label{eq:ce}
\end{align}
\end{minipage}
In fact, DiGress solely uses $\Ls_t^{CE}(\theta) $  to train their diffusion model. In this paper, we adopt a hybrid loss \cite{D3PM}, that is $\Ls_t(\theta) + \lambda \Ls_t^{CE}(\theta) $ with $\lambda =0.1$ at each time $t$, as we found it to help reduce overfitting. 
To generate a graph from $p_\theta(\ermG_0)$, a pure noise graph is first sampled from $p_\theta(\ermG_T)$ and gradually denoised using the learned $p_\theta(\ermG_{t-1}|\ermG_t)$ from step $T$ to 0. 

A significant advantage of diffusion models is their ability to achieve exchangeable probability in combination with permutation equivariant networks under certain conditions \cite{xu2022geodiff}.  
DiGress is the first work that applied discrete denoising diffusion to graph generation, achieving significant improvement over previous continuous-state based diffusion. 
However, given the inherently high-dimensional nature of graphs and their complex internal dependencies, modeling the joint distribution of all nodes and edges directly presents significant challenges. DiGress requires thousands of denoising steps to accurately capture the original dependencies. Moreover, DiGress relies on many extra supplementary node and graph-level features, such as cycle counts and eigenvectors, to effectively break symmetries
{among structural equivalences} to achieve high performance.

\subsection{Autoregressive Graph Generation} \label{ssec:autoregressive}

Order is important for AR models.
Unlike diffusion models that aim to capture the joint distribution directly, AR models decompose the joint probability into a product of simpler conditional probabilities based on an order. This makes AR models inherently suitable for ordinal data, where a natural order exists, such as in natural languages and images.

\textbf{Order Sensitivity}. 
Early works of graph generation contain many AR models like GraphRNN \cite{GraphRNN} and GRAN \cite{GRAN} based on non-deterministic heuristic  node orders like BFS/DFS and k-core ordering. Despite being permutation sensitive, AR models achieve SOTA performance on small simulated structures like grid and lobster graphs. However, permutation invariance is necessary for estimating an accurate likelihood of a graph, and can benefit large-size datasets for better generalization. 

Let $\pi$ denote an ordering of nodes. 
To make AR order-\textit{insensitive}, there are two directions: (1) Modeling the joint probability $p(\ermG, \pi)$ and then marginalizing $\pi$, (2) Finding a unique canonical order $\pi^*(\ermG)$ for any graph $\ermG$ such that $p(\pi|\ermG)$ = 1 if $\pi=\pi^*(\ermG)$ and 0 otherwise. In  direction (1), directly integrating out $\pi$ is prohibitive as the number of permutations is factorial in the graph size. Several studies \cite{li2018learning, order_matters,GRAN} have used subsets of either random or canonical orderings. This approach aims to simplify the process, but it results in approximated integrals with indeterminate errors. Moreover, it escalates computational expense due to the need for data augmentation involving these subsets of orderings. In direction (2), identifying a universal canonical order for all graphs is referred to as graph canonicalization.  
There exists no polynomial time solution for this task on general graphs, which is at least as challenging as the NP-intermediate Graph Isomorphism problem \cite{arvind2007space}. 
%While graph canonicalization has not been shown in general to be intractable (NP-complete), no polynomial general solutions are known, except for restricted classes of graphs such as molecules \cite{carbonell2013stereo}.
\citet{graphgen} explored using minimum DFS code to construct canonical labels for a specific dataset with non-polynomial time complexity. However, the canonicalization is specific to each training dataset with the randomness derived from DFS.  This results in a generalization issue, due to the canonical order being $\pi(\ermG|\text{TrainSet})$ instead of $\pi(\ermG)$. 

\textbf{The Existence of Partial Order}.
While finding a unique order for all nodes of a graph is NP-intermediate, we argue that finding a unique \textit{partial} order, where certain nodes and edges are with the same rank, is easily achievable. For example, a trivial partial order is simply all nodes and edges having the same rank. Nevertheless, a graph is not the same as a set (a set is just a graph with empty $\gE$), where all elements are essentially unordered with equivalent rank. That is because a non-empty graph contains edges between nodes, and these edges give  different structural properties to nodes. Notice that some nodes or edges have the same structural property as they are structurally equivalent. We can view each structural property as a color, and rank all unique colors within the graph to define the partial order over nodes, which we call a \textit{structural partial order}. The structural partial order defines a sequence of \textit{blocks} such that all nodes within a block have the same rank (i.e. color). 

Let $\phi:\gV \rightarrow [1,...,K_B]$ be the function that assigns rank to nodes based on their structural properties, where $K_B$ denotes the maximum number of blocks.  We use $G[\gS]$ to denote the induced subgraph on the subset $\gS \subseteq \gV$. There are many ways to assign rank to structural colors, however we would like the resulting partial order to satisfy certain constraints. Most importantly, we want 
\begin{align}
    \forall r \in [1,...,K_B],\  G[\phi(\gV)\leq r ] \text{ is a connected graph.}
\end{align}

The connectivity requirement is to ensure a more accurate representation of real-world graph generation processes, where it is typical of real-world dynamic graphs to  enlarge with newcoming nodes being connected at any time. Then, \textit{one can sequentially remove all nodes with the lowest degree to maintain this connectivity and establish a partial order.} However, degree only reflects limited information of the  first-hop neighbors, 
and many nodes share the same degree---leading to only a few distinct blocks, not significantly different from a trivial, single-block approach.

\begin{figure}[t]
\vspace{-0.3in}
    \centering
    \begin{minipage}{0.54\textwidth}
        \centering
        % Algorithm
        \input{Algorithms/partial_order}
    \end{minipage}
    \hfill
    \begin{minipage}{0.44\textwidth}
        \centering
        % Figure
        \includegraphics[width=\textwidth]{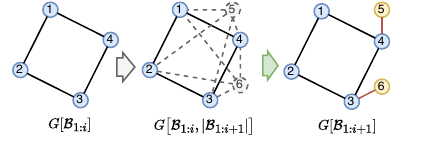}
        \caption{Example case  where the equivariant graph transformation from $G[\gB_{1:i}]$ to $G[\gB_{1:i+1}]$ is impossible for \textit{any} permutation-equivariant network due to structural equivalence of nodes. See Proposition \ref{prop:gt}.}\label{fig:flaw}
    \end{minipage}
    \vspace{-0.2in}
\end{figure}

To ensure connectivity while reducing rank collision, we consider larger hops to define a weighted degree. Consider a maximum of $K_h$ hops. For any node $\vv \in \gV$, the number of neighbors at each hop of $\vv$ can be easily obtained as $[d_1(\vv),..., d_{K_h}(\vv)]$. We then define the weighted degree as 
\begin{align}
% \vspace{-0.1in}
    w_{K_h}(\vv) =  \sum_{k=1}^{K_h} d_k(\vv) \times |\gV|^{K_h - k} \label{eq:wd}
    % \vspace{-0.1in}
\end{align}
Equation \eqref{eq:wd} may appear as an ad-hoc design, but it fundamentally serves as a hash that maps the vector $[d_1(\boldsymbol{v}), d_2(\boldsymbol{v}),..., d_K(\boldsymbol{v})]$ to a unique scalar value. This mapping ensures a one-to-one correspondence between the vector and the scalar. Furthermore, it prioritizes lower-hop degrees over higher-hop degrees, akin to how e.g. the number "123" is represented as $1 \times 10^2 + 2 \times 10^1 + 3 \times 10^0$. \eqref{eq:wd} is efficient to compute and guarantees: 1) Nodes have the same rank if and only if they have the same number of neighbors up to $K_h$ hops. 2) Lower-hop degrees are weighted more heavily.
With $w_{K_h}$ defined, we present our structural partial order in Algo. \ref{alg:partial_order}.

% Equation \eqref{eq:wd} is efficient to compute and guarantees that:
% 1) Nodes have the same rank if and only if they have the same number of neighbors up to $K_h$ hops.
% 2) Lower-hop degrees are weighted more heavily, so nodes with smaller lower-hop degrees have smaller $w_{K_h}$ values.
% With $w_{K_h}$ defined, we present the structural partial order in Algo. \ref{alg:partial_order}.

\begin{proposition}\label{eq:order-equivariant}
For any $G$, its structural partial order $\phi$ defined by Algo. \ref{alg:partial_order} is permutation equivariant, such that $\phi(\mP \star \ermG ) = \mP \star \phi(\ermG)$ for any permutation operator $\mP$. 
\vspace{-0.1in}
\end{proposition}

It is easy to prove Prop. \ref{eq:order-equivariant}. Algo. \ref{alg:partial_order} shows that $\phi(i)$ for $\forall i \in \gV$ is uniquely determined by node $i$'s structural higher-order degree. As nodes' higher-order degree is permutation equivariant, $\phi$ is also permutation equivariant. Notice that $\phi$ is unique and deterministic for any graph.

\textbf{Autoregressive Blockwise Generation.} $\phi$  in Algo. \ref{alg:partial_order} with output in range $[1,K_B]$ divides the nodes  $\gV(G)$ into $K_B$ blocks $[\gB_1,...,\gB_{K_B}]$ in order, where $\gB_{j} = \{ i\in \gV(G) | \phi(i) = j \}$. Let $\gB_{1:i}:= \cup_{j=1}^i \gB_j$ be the union of the first $i$ blocks. 
% Notice that $\gB_i$ is a set but can also be represented as a mask $\in [0,1]^{|\gV|}$ where its $j$-th element being 1 indicates node $j$ is in $\gB_i$. 
\method decomposes the joint probability of a graph $G$ into 
\begin{align}
    p_\theta(\ermG) = \prod_{i=1}^{K_B} p_\theta\Big(\ermG[\gB_{1:i}] \setminus \ermG[\gB_{1:i-1}] \  \Big|\  \ermG[\gB_{1:i-1}] \Big) \label{eq:p_conditional}
\end{align}
where $\ermG[\gB_{1:0}]$ is defined as the empty graph, and $\ermG[\gB_{1:i}] \setminus \ermG[\gB_{1:i-1}]$ denotes the set of nodes and edges that are present in $\ermG[\gB_{1:i}]$ but not in $\ermG[\gB_{1:i-1}]$. All of them are represented in natural order of $G$. As each conditional probability only contains a subset of edges and nodes, and having access to all previous blocks, this conditional probability is significantly easier to model than the whole joint probability. Given the property Prop. \ref{eq:order-equivariant} of $\gB_i$, it is easy to verify that $p_\theta(\ermG)$ is exchangeable with permutation-invariant probability for any $G$ if and only if all conditional probabilities are exchangeable. See Appx. \S\ref{apdx:exchangeable} for details of permutation-invariant probability. Note that while GRAN \cite{GRAN} also generates graphs block-by-block, all nodes within GRAN have different generation ordering, even within the same block (it breaks symmetry for the problem identified later). Essentially, GRAN is an autoregressive method and, as such, suffers from all the disadvantages inherent to AR.
% : sensitive to generation ordering, need more training epochs for data augmentation of different generation ordering, and probability estimation changes with permutation.  

\subsection{Impossibility of Equivariant Graph Transformation}\label{ssec:annealing}
\vspace{-0.1in}

In \eqref{eq:p_conditional}, we need to parameterize the conditional probability $p_\theta\Big(\ermG[\gB_{1:i}] \setminus \ermG[\gB_{1:i-1}] \  \Big|\  \ermG[\gB_{1:i-1}] \Big)$ to  be permutation-invariant. This can be achieved by letting the conditional probability be 
\begin{align}
p_\theta\Big(|\gB_i|\ \Big| \  \ermG[\gB_{1:i-1}]  \Big)  \prod_{ \substack{\rvx \in \ermG[\gB_{1:i}] \setminus  \ermG[\gB_{1:i-1}] }}  p_\theta\Big(\rvx \Big|\  \ermG[\gB_{1:i-1}]  \cup \emptyset[\gB_{1:i}] \Big)  \label{eq:p_conditional_decom}
\end{align}
where $\rvx$ is any node and edge in $\ermG[\gB_{1:i}] \setminus \ermG[\gB_{1:i-1}]$,  $\emptyset$ denotes an empty graph, hence  $\ermG[\gB_{1:i-1}]  \cup \emptyset[\gB_{1:i}]$ depicts augmenting  $\ermG[\gB_{1:i-1}] $ with empty (or virtual) nodes and edges to the same size as $\ermG[\gB_{1:i}] $. 
With the augmented graph, we can parameterize $p_\theta\big(\rvx \ \big|\  \ermG[\gB_{1:i-1}]  \cup \emptyset[\gB_{1:i}] \big)$ for any node and edge $\rvx$ with a permutation equivariant network to achieve the required permutation invariance. For  simplicity, let  $\ermG\big[\gB_{1:i-1}$, $|\gB_i|\big]$ $:=\ermG[\gB_{1:i-1}]  \cup \emptyset[\gB_{1:i}]$. 

\textbf{The Flaw in Equivariant Modeling}. Although the parameterization in \eqref{eq:p_conditional_decom} along with an equivariant network makes the conditional probability in \eqref{eq:p_conditional} become permutation-invariant, 
we have found that the equivariant graph transformation $p_\theta(\rvx \ | \ \ermG\big[\gB_{1:i-1}, |\gB_i|\big]) $ cannot be achieved in general for \textit{any} permutation equivariant network, no matter how powerful it is (!) For definition of graph transformation, see Appx.\S\ref{apdx:definition}. The underlying cause is the symmetry of structural equivalence, which is also a problem in link prediction \cite{Srinivasan2020On, zhang2021labeling}. Formally, let $\mA(G) $ be the adjacency matrix  of $G$ (ignoring labels) based on $G$'s default node order, then an \textit{automorphism} $\sigma$ of $G$ satisfies 
\begin{align}
    \mA(G) = \mA(\sigma \star G)
\end{align}
where $\sigma \star G$ is a reordering of nodes based on the mapping $\sigma$. Then the automorphism group  is  
\begin{align}
    \text{Aut} (G) = \{ \sigma \in \sP_{|\gV|} \ | \  \mA(G) = \mA(\sigma \star G) \}
\end{align}
where $\sP_n$ denotes all permutation mappings for size $n$. 
That is, $\text{Aut}(G)$ contains all automorphisms of $G$.  For a node $i$ of $G$, the \textit{orbit} that contains node $i$ is defined as
\begin{align}
    o(i) = \{ \sigma(i) \ | \ \forall \sigma \in \text{Aut}(G)\}\;.
\end{align}
In words, the orbit $o(i)$ contains all nodes that are \textit{structurally equivalent} to node $i$ in $G$. Two edges $(i,j)$ and $(u,v)$ are structurally equivalent if $\exists \sigma \in \text{Aut}(G)$, such that $\sigma(i)=u$ and $\sigma(j) = v$. 

\begin{theorem} \label{thm:structural-equivalence}
Any structurally equivalent nodes and edges will have identical representations in any equivariant network, regardless of its power or expressiveness.
\vspace{-0.05in}
\end{theorem} 
See proof in Apdx.\S\ref{apdx:proof-structural-equivalence}. Theorem \ref{thm:structural-equivalence} indicates that no matter how powerful the equivariant network is, any structually equivalent elements have the same representation, which implies the following.

\begin{proposition}
\label{prop:gt}
    General graph transformation is not achievable with any equivariant model. 
    \vspace{-0.1in}
\end{proposition}
\begin{proof}
    To prove it, we only need to show there are many ``bottleneck'' cases where the transformation cannot be achieved. Fig.  \ref{fig:flaw} shows a case where $G[\gB_{1:}]$ is a 4-cycle, and the  next target block contains two additional nodes, each with a single edge connecting to one of the nodes of $G[\gB_{1:}]$. It is easy to see that nodes $1$--$4$ are all structurally equivalent, and so are nodes $5, 6$ in the augmented case (middle). Hence, edges in $\{(5,i) | \forall i \in [1,4]\}$ are structurally equivalent (also $\{(6,i) | \forall i \in [1,4]\}$). Similarly, $\forall i \in [1,4],$ edge $(5,i)$ and $(6,i)$ are structurally equivalent. Combining all cases, edges in $\{(j,i) | \forall i \in [1,4], j \in \{5,6\}\}$ are structurally equivalent. Theorem \ref{thm:structural-equivalence} states that all these edges would have the \textit{same} prediction, hence making the target $G[\gB_{1:i+1}]$ not achievable. 
\end{proof}

\subsection{\method: Autoregressive Denoising Diffusion}

\textbf{The Magic of Annealing/Randomness}. 
In Fig. \ref{fig:flaw} we showed that a graph with many automorphisms %inside 
cannot be transformed to a target graph with fewer automorphisms. We hypothesize that \textit{a graph with lower ``energy'' is hard to be transformed to a graph with higher ``energy'' with equivariant networks}. There exist some definitions and discussion of graph energy \cite{gutman2009energy, balakrishnan2004energy} based on symmetry and eigen-information to measure graph complexity, where graphs with more symmetries have lower energy. The theoretical characterization of the conditions for successful graph transformation is a valuable direction, which we leave for future work to investigate. 

Based on the above hypothesis, to achieve a successful transformation of a graph into a target graph, it is necessary to increase its energy. Since graphs with fewer symmetries exhibit higher energy levels, our approach involves adding random noise to nodes and edges. Our approach of elevating the energy level, followed by its reduction to attain desired target properties, mirrors the annealing process. 

\textbf{Diffusion.} This further motivates us to use denoising diffusion to model $p_\theta(\rvx \ | \ \ermG\big[\gB_{1:i-1}, |\gB_i|\big])$: it naturally injects noise in the forward process, and its backward denoising process is the same as annealing. As we show below,  this yields $ p_\theta(G)$ in \eqref{eq:p_conditional} to be permutation-invariant.

% What is more, we can achieve the permutation-invariant property for  $p_\theta\Big(\ermG[\gB_{1:i}] \setminus \ermG[\gB_{1:i-1}] \  \Big|\  \ermG[\gB_{1:i-1}] \Big)$, based on Proposition 1 in  \cite{xu2022geodiff}. 
%https://arxiv.org/pdf/2203.02923.pdf
% Finally, as we have analyzed, this yields $ p_\theta(G)$ in \eqref{eq:p_conditional} to be permutation-invariant. 

\begin{theorem}
    \method is permutation-invariant such that $p_\theta(\mP \star \ermG ) =  p_\theta(\ermG )$ $\forall \text{ permutator }\mP$.
\end{theorem}
The proof is given in Appx. \S\ref{apdx:exchangeable}. 
With all constituent parts presented, we 
 summarize our proposed \method, the first permutation-invariant autoregressive diffusion model that integrates AR with denoising diffusion. \method relies on a unique, permutation equivariant structural partial order $\phi$ (Algo. \ref{alg:partial_order}) to decompose the joint graph probability to the product of simpler conditional probabilities, based on  \eqref{eq:p_conditional}. Each block's conditional probability is modeled with the product of a conditional block size probability and a conditional block enlargement probability as in \eqref{eq:p_conditional_decom}, where the %conditional block enlargement probability 
 latter
 for every block is  a shared discrete denoising diffusion model as described in \S\ref{ssec:diffusion}.  Fig. \ref{fig:pard-visual} illustrates \method's two parts : (top) block-wise AR and (bottom) local denoising diffusion at each AR step.

Notice that  there are two tasks in \eqref{eq:p_conditional_decom}; one for predicting the next block's size, and the other for predicting the next block's nodes and edges with diffusion. These two tasks can be trained together with a single network, although for better performance we use two different networks. For each block's diffusion model, we set the maximum time steps to 40 without much tuning.

\input{Figures/bigpic}

\textbf{Training and Inference.} We provide the training and inference algorithms for \method in Apdx.\S\ref{apdx:train-inference}. Specifically, Algo. \ref{alg:train-blocksize} is used to train next block's size prediction model; Algo. \ref{alg:train-diffusion} is used to train the shared diffusion for block conditional probabilities; and Algo. \ref{alg:generation} presents the generation steps.

%% file: Algorithms/partial_order.tex
\begin{algorithm}[H]
\caption{Structural Partial Order $\phi$}
\small
\label{alg:partial_order}
\begin{algorithmic}[1]
\STATE {{\bfseries Input:}} Graph $G$, maximum hops $K_h$.
\STATE {{\bfseries Init:}} $G_0 = G$, $i=0$, $\phi$ with $\phi(\vv) = 0 \  \forall \vv.$
\WHILE{$G_i$ is not $\emptyset$}
\STATE Compute $w_{K_h}(\vv),\forall \vv \in \gV(G_i),$ using \eqref{eq:wd}.
\STATE Find all nodes $\cL$ with $w_{K_h} = \min_{\vv \in \gV(G)}w_{K_h}(\vv)$.
\STATE Let $\phi (\vv) = i \ \forall \vv \in \cL $.
\STATE $G_{i+1}\leftarrow G_i [\gV(G) \setminus \cL]$; $i \leftarrow i+1$.
\ENDWHILE
\STATE {{\bfseries Output:}}  $\phi \leftarrow i - \phi$
\end{algorithmic}
\end{algorithm}
\setlength{\textfloatsep}{12pt}

%% file: Figures/bigpic.tex
\begin{figure}[!t]
\vspace{-0.1in}
    \centering
    \hspace{-0.4in}
\includegraphics[width=0.6\textwidth]{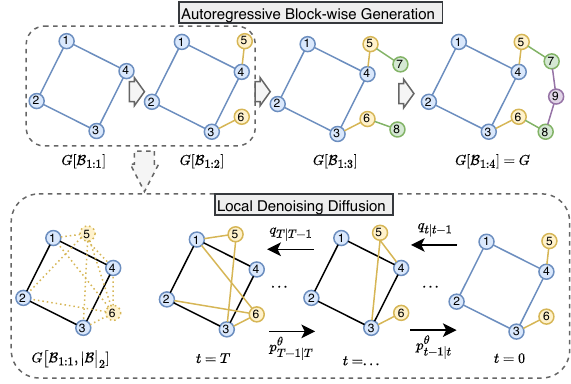}
    % \vspace{-20pt}
    % \vspace{-0.05in}
    \caption{\method integrates the autoregressive method with diffusion modeling. (top) \method decomposes the joint probability into a series of block-wise enlargements, where each block's conditional distribution is captured with a shared discrete diffusion (bottom). }
    \label{fig:pard-visual}
    \vspace{-0.3in}
\end{figure}

%% file: 4-realization.tex
\method is a general framework that can be combined with any equivariant network. Nevertheless, we would like an equivariant network with enough expressiveness to process symmetries inside the generated blocks for modeling the next block's conditional probability. While there are many expressive GNNs like subgraph GNNs \cite{bevilacqua2022equivariant,zhao2022from} and higher-order GNNs \cite{zhao2022practical, morris2022speqnets}, PPGN \cite{PPGN} is still a natural choice that models edge (2-tuple) representations directly with 3-WL expressivity and $O(n^3)$ complexity in graph size. However, PPGN's memory cost is relatively high for many datasets. 

\subsection{Efficient and Expressive Higher-order Transformer}
\label{ssec:model}

To enhance the memory efficiency of PPGN while maintaining the expressiveness equivalent to the 3-Weisfeiler-Lehman (3-WL) test, we introduce a hybrid approach that integrates Graph Transformers with PPGN. Graph Transformers operate on nodes as the fundamental units of representation, offering better scalability and reduced memory consumption ($O(n^2)$) compared to PPGN. PPGN utilizes edges as their primary representation units and therefore incurs significantly higher memory requirements ($O(n^3)$). However, the expressiveness of Graph Transformers (without position encoding) is limited to the 1-WL test \cite{cai2023connection}. By combining these two models, we can drastically decrease the size of edge representations while allocating larger hidden sizes to nodes. This synergistic approach not only substantially lowers the memory footprint but also enhances overall performance, leveraging the strengths of both architectures to achieve a balance between expressivity and efficiency. We provide the detailed design in Appx.\S\ref{apdx:ppgn-transformer}. Note that we use GRIT \cite{ma2023GraphInductiveBiases} as the graph transformer block.

% \subsection{Input Conditioning}

\subsection{Parallel Training with Causal Transformer}
As shown in \eqref{eq:p_conditional}, for a graph $G$, there are $K_B$ conditional probabilities 
% $p_\theta\Big(\ermG[\gB_{1:i}] \setminus \ermG[\gB_{1:i-1}] \  \Big|\  \ermG[\gB_{1:i-1}] \Big)$ 
being modeled by a shared diffusion model using a $\theta$-parameterized network $f_\theta$. By default, these $K_B$ number of inputs $\{\ermG[\gB_{1:i-1}]\}_{i=1}^{K_B}$ are viewed as separate graphs and the representations during network passing $f_\theta (\ermG[\gB_{1:i-1}])$ for different $i\in [1,K_B]$ are not shared.  This leads to a scalability issue; in effect enlarging the dataset by roughly $K_B$ times and resulting in $K_B$ times longer training. 

To minimize computational overhead, it is crucial to enable parallel training of all the $K_B$ conditional probabilities, and allow these processes to share representations, through which we can pass the full graph $G$ to the network $f_\theta$ only once and obtain all $K_B$ conditional probabilities.
This is also a key advantage of transformers over RNNs. 
Transformers (GPTs) can train all next-token predictions simultaneously with representation sharing through causal masking, whereas RNNs must train sequentially.
However, the default causal masking of GPTs is not applicable to our architecture, as \method contains both Transformer and PPGN where the PPGN's causal masking is not designed.

To ensure representation sharing without risking information leakage, we first assign a ``block ID'' to every node and edge within graph $G$. Specifically, for every node and edge in $G[\gB_{1:i}] \setminus G[\gB_{1:i-1}]$, we assign the ID equal to $i$. To prevent information leakage effectively, it is crucial that any node and edge labeled with ID $i$ are restricted to communicate only with other nodes and edges whose ID is $\leq i$. Let $\mA ,\mB \in \sR^{n\times n}$, and $\mathbf{x} \in \sR^{n}$.
There are mainly two non-elementwise operations in Transformer and PPGN that have the risk of leakage: the attention-vector product operation $\mA \mathbf{x}$ of Transformer, and the matrix-matrix product operation $\mA\mB$ of PPGN. (We ignore the feature dimension of $\mA$ and $\mathbf{x}$ as it does not affect the information leakage.)                         Let $\mM \in \{0,1\}^{n\times n}$ be a mask matrix, such that $\mM_{i,j} = 1 $ if $\text{block\_ID}(\text{node } i) \ge \text{block\_ID}(\text{node }j) $ else 0. One can verify that 
\begin{align}
   (\mA \odot \mM)\mathbf{x}, \quad \text{and} \quad
   (\mA \odot \mM)\mB + \mA (\mB\odot \mM^\top) - (\mA \odot \mM)(\mB\odot \mM^\top) \label{eq:causal_matrix_prod}
\end{align}
 generalize  $\mA \mathbf{x}$ and $\mA \mB$ respectively and safely bypass information leakage. We provide more details of parallel training and derivation of \eqref{eq:causal_matrix_prod} in Appx.\S\ref{apdx:causal}. 
We use these operations in our network and enable representation sharing, along with parallel training of all $K_B$ blocks  for denoising diffusion as well as next block size prediction. In practice, these offer more than 10$\times$ speed-up, and the parallel training allows \method to scale to large datasets like MOSES \cite{moses}.

%% file: 5-experiments.tex
% add ablation study 

We evaluate \method on 8 diverse benchmark datasets with varying sizes and structural properties, including both molecular (\S\ref{ssec:molecule}) and non-molecular/generic (\S\ref{ssec:generic}) graph generation. 
A summary of the datasets and details are in Appx. \S\ref{ssec:exp_details}. Ablations and runtime measures are in Appx. \S\ref{apdx:ablation}. 

\subsection{Molecular Graph Generation}
\label{ssec:molecule}

{\bf Datasets.} We experiment with three different molecular datasets used across the graph generation literature: 
(1) \qm \cite{ramakrishnan2014quantum}
(2) \zinc \cite{irwin2012zinc}, and 
(3) \moses \cite{moses} that contains more than 1.9 million graphs. 
We use a 80\%-20\% train and test split, and among the train data we split additional 20\% as validation.
For \qm and \zinc, we generate 10,000 molecules for stand-alone evaluation, and on \moses we generate 25,000 molecules.

{\bf Baselines.~} The literature has not been consistent in evaluating molecule generation on well-adopted benchmark datasets and metrics. Among baselines,
DiGress \cite{DiGress} stands out as the most competitive. We also compare to many other baselines shown in tables, such as GDSS \cite{GDSS} and GraphARM \cite{GraphARM}.

{\bf Metrics.~}
The literature has adopted a number of different evaluation metrics that are not consistent across datasets. 
Most common ones include
Validity ($\uparrow$) , Uniqueness ($\uparrow$) (frac. of valid molecules that are unique), and Novelty ($\uparrow$) (frac. of valid molecules that are not included in the training set).
For \qm, following earlier work \cite{DiGress}, we report additional evaluations w.r.t. Atom Stability ($\uparrow$) and Molecule Stability ($\uparrow$), as defined by \cite{hoogeboom22a}, whereas Novelty is not reported as explained in \cite{DiGress}.
On \zinc and \moses, we also measure the Fr\'{e}chet ChemNet Distance (FCD) ($\downarrow$) between the generated and the training samples, which is based on the embedding learned by ChemNet \cite{li2018learning}. For \moses, there are three additional measures: Filter ($\uparrow$) score 
is the fraction of molecules passing the same filters as the test set,  SNN ($\uparrow$)  evaluates nearest neighbor similarity using Tanimoto Distance, and Scaffold similarity ($\uparrow$) analyzes the occurrence of Bemis-Murcko scaffolds \cite{moses}.

\input{Tables/qm9-zinc}
{\bf Results.~}
Table \ref{tab:qm9_result} shows generation evaluation results on \qm, where the baseline results are sourced from \cite{DiGress}. 
\method outperforms DiGress and variants that do {\em not} use any auxiliary features, with slightly lower Uniqueness. 
What is notable is that \method, without using any extra features, achieves a similar performance gap  against 
DiGress that uses  specialized extra features. Table \ref{tab:zinc_result} shows \method's performance on \zinc, with baseline results carried over from \cite{GraphARM} and \cite{SwinGNN}. \method achieves the best Uniqueness, stands out in FCD alongside SwinGNN \cite{SwinGNN}, and is the runner-up w.r.t. Validity.  

\input{Tables/moses}

Finally, Table \ref{table:moses} shows generation quality 
on the largest dataset \moses. We mainly compare with  DiGress and its variants,  which has been the only general-purpose generative model in the literature that is not based on molecular fragments or SMILES strings. %While some of the baselines achieve high validity, they involve hard-coded rules to ensure generative quality.
Baselines are sourced from \cite{DiGress}. While other specialized models, excluding \method and DiGress, have hard-coded rules to ensure high Validity, \method outperforms those on several other metrics including FCD and SNN, and achives competitive performance on others.
Again, it is notable here that \method, \textit{without} relying on any auxiliary features, achieves similarly competitive results as with 
DiGress which utilizes extra features.

\subsection{Generic Graph Generation}
\label{ssec:generic}

\input{Tables/generic_graph}

{\bf Datasets.~} We use five generic graph datasets with various structure and semantic:  
(1) \comm \cite{GraphRNN}, 
(2) \cave \cite{caveman}, 
(3) \cora \cite{sen2008collective},   
(4) \breast \cite{gonzalez2011high}, and
(5) \grid \cite{GraphRNN}.
%(5) \enz \cite{schomburg2004brenda} and 
%(6) \ego \cite{sen2008collective}.
We split each dataset into 80\%-20\% train-test, and randomly sample 20\% of training graphs for validation. 
We generate the same number of samples as the test set. Notice that \grid contains graphs with 100$\sim$400 nodes, which is relatively large for diffusion models. There are lots of symmetries inside, hence it is difficult to capture all dependencies with permutation-equivariant models.

{\bf Baselines.~} We mainly compare against the latest general-purpose GraphArm \cite{GraphARM}, which reported DiGress \cite{DiGress} and GDSS \cite{GDSS} as top two most competitive, along with several other baselines.

{\bf Metrics.~} We follow \cite{GraphRNN} to measure generation quality using the maximum mean discrepancy (MMD) as a distribution distance  between the generated graphs and the test graphs ($\downarrow$), as pertain to distributions of ($i$) Degree, ($ii$) Clustering coefficient, and ($iii$) occurrence count of all Orbits.

{\bf Results.}
Table \ref{tab:generic} provides the generation results of \method against the baselines as sourced from \cite{GraphARM}. \method shows outstanding performance achieving SOTA or close {runner-up} results, while none of the baselines shows as consistent performance across datasets and metrics.

\subsection{Ablation Study}
\label{ssec:ablation}

\textbf{Q1: do we really need AR in diffusion, given diffusion is already permutation-invariant?}

In DiGress \cite{DiGress}, we observed that pure diffusion, while being permutation-invariant, requires (1) many sampling/denoising steps to break symmetries and (2) additional features like eigenvectors to further break symmetry. This shows that directly capturing the FULL joint distribution and solving the transformation difficulty (in Sec. \ref{ssec:annealing}) via diffusion is challenging. Additionally, AR methods still dominate LLMs, indicating their potential to benefit diffusion models. To quantitatively verify our analysis, we perform an ablation study on the maximum number of hops, $K_h$, which controls the extent of autoregression (AR). When $K_h=0$, all nodes have a fixed degree of 1, resulting in a single block per graph, equivalent to full diffusion without AR. As $K_h$ increases, more blocks are generated with smaller average block sizes, indicating a greater number of AR steps.

\input{Tables/hops}

Table \ref{tab:hops_ablation} shows the result of the controlled experiment with 140 total diffusion steps across all trials, using the same model architecture, diffusion algorithm, and training settings. The significant improvement from $K_h=0$ to $K_h=1$
 confirms that our enhancement stems from breaking the full joint distribution into several conditional distributions. Furthermore, adding more diffusion steps to full diffusion approach does not close the gap to AR enhanced diffusion, indicates the necessary of combining AR and diffusion. 

\textbf{Q2: how does model architecture affect performance?}

\input{Tables/arch}
Table \ref{tab:arch_ablation} this indicates that PPGN component is essential for diffusion with autoregression. The design of combining PPGN and transformer in Sec. \ref{sec:realization} further addresses the efficiency of PPGN.

\textbf{Other ablations:} other ablations and runtime measures are in Appx. \S\ref{apdx:ablation}

%% file: Tables/qm9-zinc.tex
\begin{table}[ht]
\vspace{-0.2in}
    \centering
    \begin{minipage}{.48\linewidth}
        \setlength{\tabcolsep}{2pt}
        \centering
        \caption{Generation quality on \qm with explict hydrogens.}
        \label{tab:qm9_result}
        \scalebox{0.85}{
        \begin{tabular}{lrrrc}
        \toprule
        \textbf{Model} & \textbf{Valid.} $\uparrow$ & \textbf{Uni.} $\uparrow$ & \textbf{Atom.}$\uparrow$ & \textbf{Mol.} $\uparrow$ \\
        \midrule
        Dataset (optimal) & 97.8 & 100 & 98.5 & 87.0 \\
        \midrule
        ConGress & 86.7  & \textbf{98.4}  & 97.2  & 69.5  \\
        DiGress (uniform) & 89.8  & 97.8  & 97.3  & 70.5  \\
        DiGress (marginal) & 92.3  & 97.9  & 97.3  & 66.8  \\
        DiGress (marg. \textit{+ feat.}) & 95.4  & 97.6 & 98.1  & 79.8  \\
        \midrule
        \method (\textit{no feat.}) &\textbf{97.5}	&95.8	&\textbf{98.4}	&\textbf{86.1}\\
        \bottomrule
        \end{tabular}
        }
    \end{minipage}%
    \hspace{0.04\linewidth}
    \begin{minipage}{.47\linewidth}
    \setlength{\tabcolsep}{2pt}
        \caption{Generation quality on \zinc.}
        \label{tab:zinc_result}
        \scalebox{0.83}{
        \begin{tabular}{lrrrr}
        \toprule
        {\textbf{Model}} & {\textbf{Validity} $\uparrow$}  & {\textbf{FCD} $\downarrow$} & {\textbf{Uni.} $\uparrow$} & \textbf{Model Size} \\
        \midrule
        EDP-GNN   & 82.97  & 16.74 & 99.79 & 0.09M  \\
        GraphEBM  & 5.29   & 35.47 & 98.79  &-  \\
        SPECTRE   & 90.20  & 18.44 & 67.05 & -   \\
        GDSS & \textbf{97.01}  & 14.66 & 99.64 & 0.37M  \\
        GraphArm  & 88.23 & 16.26 & 99.46 & - \\
        DiGress  & 91.02  & 23.06 & 81.23 & 18.43M   \\
        SwinGNN-L & 90.68 & 1.99 & 99.73 & 35.91M\\
        \midrule
        \method &  95.23 &	\textbf{1.98} & \textbf{99.99}& 4.1M	\\
        \bottomrule
        \end{tabular}
        }
    \end{minipage}
    \vspace{-0.2in}
\end{table}

%% file: Tables/moses.tex
\begin{table}[h]
% \vspace{-0.4in}
\footnotesize
\setlength{\tabcolsep}{3.0pt}
\centering
\caption{Generation quality on \moses. The top three methods use hard-coded rules (not highlight).}
\label{table:moses}
\scalebox{0.9}{
\begin{tabular}{lccccccc}
\toprule
\textbf{Model}      & \textbf{Val.} \textuparrow & \textbf{Uni.} \textuparrow & \textbf{Novel.} \textuparrow & \textbf{Filters} \textuparrow & \textbf{FCD} \textdownarrow & \textbf{SNN} \textuparrow & \textbf{Scaf.} \textuparrow \\
\midrule
VAE      & 97.7 & 99.8 & 69.5 & 99.7 & 0.57 & 0.58 & 5.9  \\
JT-VAE         & 100 & 100 & 99.9 & 97.8 & 1.00 & 0.53 & 10.0   \\
\scalebox{0.7}{GraphINVENT}    & 96.4 & 99.8 & -    & 95.0 & 1.22 & 0.54 & 12.7 \\
\midrule
ConGress   & 83.4 & 99.9 & \textbf{96.4} & 94.8 & 1.48 & 0.50 & \textbf{16.4} \\
DiGress   & 85.7 & \textbf{100}  & 95.0 & 97.1 & 1.19 & 0.52 & 14.8 \\
\midrule
\method &\textbf{ 86.8}  &	\textbf{100}&	78.2&	\textbf{99.0}	&\textbf{1.00} &	\textbf{0.56}&	2.2\\
\bottomrule
\end{tabular}
}
% \vspace{-0.2in}
\end{table}

%% file: Tables/generic_graph.tex
\begin{table*}[ht]
\vspace{0in}
\setlength{\tabcolsep}{2.5pt}
\centering
\footnotesize
\vspace{-0.05in}
\caption{Generation quality on generic graphs. All metrics are based on generated-to-test set MMD distances, the lower the better. Top performance is in \textbf{bold}, and Runner-up is \underline{underlined}.}
\vspace{-0.1in}
\scalebox{0.87}{
\begin{tabular}{l|ccc|ccc|ccc|ccc|ccc}
\toprule
 & \multicolumn{3}{c|}{\textbf{\comm}} & \multicolumn{3}{c|}{\textbf{\cave}} & \multicolumn{3}{c|}{\textbf{\cora}} & \multicolumn{3}{c}{\textbf{\breast}} &  \multicolumn{3}{c}{\textbf{\grid}} \\ 
\midrule
\multicolumn{1}{c|}{\textbf{Model}}& \textbf{Deg.} & \textbf{Clus.} & \textbf{Orbit} & \textbf{Deg.} & \textbf{Clus.} & \textbf{Orbit} & \textbf{Deg.} & \textbf{Clus.} & \textbf{Orbit} & \textbf{Deg.} & \textbf{Clus.} & \textbf{Orbit} & \textbf{Deg.} & \textbf{Clus.} & \textbf{Orbit}\\ 
\midrule
GraphRNN  & 0.080 & 0.120 & 0.040 & 0.371 & 1.035 & 0.033 &  1.689 & 0.608 &0.308   & 0.103 & 0.138 & 0.005 & 0.064 & 0.043 & 0.021\\
GRAN      & 0.060 & 0.110 & 0.050 & 0.043 & 0.130 & 0.018 & 0.125 & 0.272 & 0.127 & 0.073 & 0.413 & 0.010 & - & - & - \\
EDP-GNN   & 0.053 & 0.144 & 0.026 & 0.032 & 0.168 & 0.030 & 0.093 & 0.269 & \underline{0.062} & 0.131 & 0.038 & 0.019 & 0.455 & 0.238 & 0.328 \\
GDSS      & 0.045 & 0.086 & \underline{0.007} & \underline{0.019} & 0.048 & 0.006 & 0.160 & 0.376 & 0.187 & 0.113 & \textbf{0.020} & 0.003 & 0.111 & 0.005 & 0.070 \\
GraphArm  & \underline{0.034} & 0.082 & \textbf{0.004} & 0.039 &\textbf{0.028} &0.018 & 0.273  & 0.138 & 0.105& \textbf{0.036} & 0.041 & \underline{0.002} & - & - & - \\
DiGress & 0.047 & \textbf{0.041} & 0.026 &  \underline{0.019}  & \underline{0.040} &\underline{0.003} & \underline{0.044} & \underline{0.042}& 0.223& 0.152 & \underline{0.024}&  0.008 & - & - & -\\
\midrule
\method & \textbf{0.023} & 
\underline{0.071} &
0.012  & \textbf{0.002}  &
0.047 &
\textbf{0.00003} & \textbf{0.0003} & \textbf{0.003} & \textbf{0.0097} & \underline{0.044} & 
\underline{0.024}
&
\textbf{0.0003}
& \textbf{0.028} & \textbf{0.002} & \textbf{0.029}\\
\bottomrule
\end{tabular}}
% \vspace{-0.15in}
\label{tab:generic}
\vspace{-0.05in}
\end{table*}

%% file: Tables/hops.tex
\begin{table}[ht]
\centering
\vspace{-0.1in}
\caption{Ablation study on QM9 with varying maximum hops while keeping the total diffusion steps fixed (first two parts). The last part examines the effect of increasing steps for the no AR case.}
\label{tab:hops_ablation}
\begin{tabular}{l|c|ccc||cc}
\toprule
\textbf{Setting}  & \textbf{No AR} & \multicolumn{3}{c||}{\textbf{With AR}} & \multicolumn{2}{|c}{\textbf{No AR, $\uparrow$ steps}} \\
\midrule
Total diffusion steps  & 140   & \multicolumn{3}{c||}{140} & 280 & 490 \\
Maximum hops          & 0     & 1     & 2     & 3   &  0 & 0  \\
Average number of blocks & 1     & 4.3   & 5.6   & 7.75 & 1  &  1 \\
Diffusion steps per block & 140   & 32    & 25    & 20   & 280 & 490\\
\midrule
Validity              & 93.8  &  \textbf{97.1}  & 96.7  & 97.0  & 94.3 & 95.2\\
Uniqueness            &  \textbf{96.9}  & 96.5  & 96.2  & 96.1 & 96.5 & 96.9 \\
Mol stability          & 76.4  & 86.1  & 85.4  &  \textbf{86.3} & 79.3 & 79.2 \\
Atom Stability        & 97.7  & 98.3  & 98.3  &  \textbf{98.4} & 97.9 & 98.0 \\
\bottomrule
\end{tabular}
\end{table}

%% file: Tables/arch.tex
\begin{table}[ht]
\centering
\vspace{-0.1in}
\caption{Results for QM9 dataset with different model architectures with $K_h=3$ and 140 total steps.}
\label{tab:arch_ablation}
\begin{tabular}{lccc}
\toprule
\textbf{Backbone}                  & \textbf{Transformer} & \textbf{PPGN} & \textbf{PPGNTransformer} \\
\midrule
Validity             & 26.3  & 96.6  & 97.1  \\
Uniqueness           & 94.5  & 96.3  & 96.0  \\
Mol Stability         & 17.6  & 84.67 & 86.2  \\
Atom Stability        & 81.4  & 98.2  & 98.4  \\
\bottomrule
\end{tabular}
\end{table}

%% file: 6-conclusion.tex
We presented \method, the \textit{first} permutation-invariant autoregressive diffusion model for graph generation.
\method decomposes the joint probability of a graph autoregressively into the product of several block conditional probabilities, by relying on a unique and permutation equivariant structural partial order. All conditional probabilities are then modeled with a shared discrete diffusion. \method can be trained in parallel on all blocks, and efficiently scales to millions of graphs. \method achieves SOTA performance on molecular and non-molecular datasets without using any extra features. Notably, we expect \method to serve as a cornerstone toward generative foundation modeling for graphs.

%% file: 8-appendix.tex
\subsection{Variational Lower Bound Derivation}
\label{ssec:vlbderive}

\begin{align}
    \log &\int q(\ermG_{1:T}|\ermG_0) \frac{p_{\theta}(\ermG_{0:T})}{q(\ermG_{1:T}|\ermG_0)}d\ermG_{1:T} 
    \geq  \E_{q(\ermG_{1:T}|\ermG_0)} \big[  \log p_{\theta}(\ermG_{0:T}) - \log q(\ermG_{1:T}|\ermG_0) \big] \nonumber\\
    &= \E_{q(\ermG_{1:T}|\ermG_0)} \big[
        \log p_{\theta}(\ermG_{0:T}) + \sum_{t=1}^T \log \frac{p_{\theta}(\ermG_{t-1}| \ermG_{t} )}{q(\ermG_{t}| \ermG_{t-1} )} 
    \big] \nonumber\\
    & = \E_{q(\ermG_{1:T}|\rvx_0)} \big[
        \log p_{\theta}(\ermG_T) + \log \frac{p_{\theta}(\ermG_{0}| \ermG_{1} )}{q(\ermG_{1}| \ermG_{0} )} +
        \sum_{t=2}^T \log \frac{p_{\theta}(\ermG_{t-1}| \ermG_{t} )}{q(\ermG_{t}| \ermG_{t-1}, \ermG_0 )} 
    \big] \nonumber\\
    & = \E_{q(\ermG_{1:T}|\ermG_0)} \big[
        \log p_{\theta}(\ermG_T) + \log \frac{p_{\theta}(\ermG_{0}| \ermG_{1} )}{q(\ermG_{1}| \ermG_{0} )} +
        \sum_{t=2}^T \log 
        \Bigg (\frac{p_{\theta}(\ermG_{t-1}| \ermG_{t} )}{q(\ermG_{t-1}| \ermG_{t}, \ermG_0 )} \cdot 
        \frac{q(\ermG_{t-1}| \ermG_0 )}{q(\ermG_{t}| \ermG_0 ) }   \Bigg )
    \big] \nonumber\\
    & = \E_{q(\ermG_{1:T}|\ermG_0)} \big[ 
        \log p_{\theta}(\ermG_T) + \log \frac{p_{\theta}(\ermG_{0}| \ermG_{1} )}{q(\ermG_{1}| \ermG_{0} )} + \log \frac{q(\ermG_1 | \ermG_0)}{q(\ermG_T | \ermG_0)} +
        \sum_{t=2}^T \log \frac{p_{\theta}(\ermG_{t-1}| \ermG_{t} )}{q(\ermG_{t-1}| \ermG_{t}, \ermG_0 )}
    \big] \nonumber\\
    & = \E_{q(\ermG_{1:T}|\ermG_0)} \big[ \log p_{\theta}(\ermG_{0}| \ermG_{1} ) + \log \frac{p_{\theta}(\ermG_T)}{q(\ermG_T | \ermG_0)} -
        \sum_{t=2}^T \log \frac{q(\ermG_{t-1}| \ermG_{t}, \ermG_0 )}{p_{\theta}(\ermG_{t-1}| \ermG_{t} )}
    \big] \nonumber\\
    &= \underbrace{\E_{q(\ermG_{1}|\ermG_0)}\big[ \log p_{\theta}(\ermG_{0}| \ermG_{1} )\big]}_{-\Ls_1(\theta)}
      - \sum_{t=2}^T \underbrace{\E_{q(\ermG_{t}|\ermG_0)}\big[ \KL\big(q(\ermG_{t-1} | \ermG_{t}, \ermG_0 ) || p_{\theta}(\ermG_{t-1}| \ermG_{t} \big)  \big]}_{\Ls_t(\theta)}
      - \text{const.}%\label{eq:vlb_1}
\end{align}

Using \eqref{eq:q}, the first term can simplified as 
\begin{align}
\E_{q(\ermG_1|\ermG_0) } [\sum_{i} \log p_\theta(\rvv^i_0|\ermG_1)
+\sum_{i,j} \log p_\theta(\rve^{i,j}_0|\ermG_1)
] \;, 
\end{align}
and similarly, the $t$-th step loss $\Ls_t(\theta)$ is 
\begin{align}
\E_{q(\ermG_t|\ermG_0)} \big[ 
&\sum_{i}  KL\big(q(\rvv^i_{t-1} | \rvv^i_{t}, \rvv^i_0 ) \ ||\   p_{\theta}(\rvv^i_{t-1}| \ermG_{t} \big)
+ \nonumber\\
&\sum_{i,j} KL\big(q(\rve^{i,j}_{t-1} | \rve^{i,j}_{t}, \rve^{i,j}_0 )\ || \ p_{\theta}(\rve^{i,j}_{t-1}| \ermG_{t} \big) 
\big] 
\end{align}

\subsection{Details of Discrete Diffusion Used in Paper}
\label{appdx:discrete-diffusion}

We closely follow the approach in \citet{zhao2024improving} to define forward and backward process in discrete diffusion, and get the formulation of three important distributions, (\textcolor{orange}{$i$}) 
 $q(\ermG_t| \ermG_0)$ (\textcolor{orange}{$ii$})  $q(\ermG_{t-1} | \ermG_t, \ermG_0)$, and (\textcolor{orange}{$iii$})  $p_{\theta}(\ermG_{t-1}| \ermG_{t} )$, needed for computing negative VLB based loss. In here, we explain the detailed constructions. Notice that we only use the most basic discrete-time discrete diffusion formulation in \citet{zhao2024improving}, mainly for improved memory efficiency over other old approaches like D3PM \cite{D3PM}. Other enhanced techniques in \cite{zhao2024improving} like approximated loss and continuous-time formulation are not used, to keep the discrete diffusion part simple.  

 DiGress \cite{DiGress} applies D3PM \cite{D3PM} to define these three terms, our approach is similar. Since all elements in the forward process are independent as shown in \eqref{eq:q}, one can verify that the two terms $q(\ermG_t | \ermG_0)$ and $q(\ermG_{t-1}|\ermG_{t}, \ermG_0)$ are in the form of a product of independent distributions on each element. 
For simplicity, we introduce the formulation for a single element $\rvx$, with $\rvx$ being $\rvv^i$ or $\rve^{i,j}$.  We assume each discrete random variable $\rvx_t$ has a categorical distribution, i.e. $\rvx_t \sim \text{Cat}(\rvx_t ; \vp)$ with $\vp\in [0,1]^{K}$ and $\1^\top \vp = 1$ . One can verify that $p(\rvx_t= \vx_t) = \vx_t^\top\vp$, or simply $p(\rvx_t) = \vx_t^\top\vp$. As shown in \citet{hoogeboom2021argmax, D3PM}, the forward process with discrete variables $q(\rvx_t | \rvx_{t-1})$ can be represented as a transition matrix  $Q_t \in [0,1]^{K\times K}$ such that $[Q_{t}]_{ij} = q(\rvx_t = \ve_j | \rvx_{t-1} = \ve_i)$. Then, 
\begin{align}
    q(\rvx_t | \rvx_{t-1}) = \text{Cat}(\rvx_t; Q^\top_t\vx_{t-1}) \;.
\end{align}
Given transition matrices $Q_1,...,Q_T$, the forward conditional marginal distribution is 
\begin{align}
  (\textcolor{orange}{i}) 
  \ q(\rvx_t | \rvx_0) = \text{Cat}(\rvx_t; \overline{Q}_{t}^\top\vx_0 ), \text{with } \overline{Q}_{t} = Q_{1}...Q_{t} \; ,\label{eq:t|0}
\end{align} 
and the $(t$$-$$1)$-step posterior distribution can be written as 
\begin{align}
(\textcolor{orange}{ii}) \ \
q(\rvx_{t-1} | \rvx_t, \rvx_0) 
&= \text{Cat}(\rvx_{t-1};\frac{Q_{t} \vx_{t} \odot \overline{Q}_{t-1}^\top\vx_0}{\vx_t^\top\overline{Q}_t^\top\vx_0 })\;.\label{eq:t-1|t,0}  
\end{align}
See Apdx.\S\ref{ssec:t-1|t,0} for the derivation.
We have the option to specify node- or edge-specific quantities, $Q_t^{v,i}$ and $Q_t^{e,{i,j}}$, respectively, or allow all nodes and edges to share a common $Q_t^v$ and $Q_t^e$. Leveraging \eqref{eq:t|0}  and \eqref{eq:t-1|t,0}, we can precisely determine $q(\rvv^i_t|\rvv^i_0)$ and $q(\rvv^i_{t-1}|\rvv^i_t,\rvv^i_0)$ for every node, and a similar approach can be applied for the edges. To ensure simplicity and a non-informative $q(\ermG_T | \ermG_0) $ (see Apdx.\S\ref{ssec:simple_trans_matrix}), we choose 
\begin{align}
    Q_t = \alpha_t I + (1-\alpha_t)\1\vm^\top
\end{align}
for all nodes and edges, where $\alpha_t \in [0, 1]$, and $\vm$ is a uniform distribution ($\1/K_v$ for nodes and $\1/K_e$ for edges). Note that DiGress \cite{DiGress} chooses $\vm$ as the marginal distribution of nodes and edges.
As $p(\rvx_{t-1}| \rvx_{t} ) =\sum_{\rvx_0}q(\rvx_{t-1} | \rvx_t, \rvx_0) p(\rvx_0 | \rvx_t)$
, the parameterization of $p_\theta (\ermG_{t-1} | \ermG_t)$ can use the relationship, with 
\begin{align}
(\textcolor{orange}{iii}) \quad p_\theta (\rvx_{t-1} | \ermG_t) = \sum_{\rvx_0}q(\rvx_{t-1} | \rvx_t, \rvx_0) p_\theta(\rvx_0 | \ermG_t) \label{eq:ptheta}
\end{align}
where $\rvx$ can be any $\rvv^i$ or $\rve^{i,j}$. With \eqref{eq:ptheta}, we can parameterize $p_\theta(\rvx_0 | \ermG)$ directly with a neural network, and compute the negative VLB loss in \eqref{eq:vlb} exactly, using Eq.s (\ref{eq:t|0}), (\ref{eq:t-1|t,0}) and (\ref{eq:ptheta}).

\subsection{Derivation of $q(x_{t-1} | x_t, x_0)$}
\label{ssec:t-1|t,0}

First, define $\overline{Q}_{t|s} = Q_{s+1}...Q_{t}$. Note that $  \overline{Q}_{t|0} = \overline{Q}_t$ and $\overline{Q}_{t|t-1} = Q_t$. Accordingly, we  can derive the following two equalities.
\begin{align}
    q(\rvx_t | \rvx_{t-1}) = \text{Cat}(\rvx_t; \overline{Q}_{t}^\top\rvx_{t-1} ) \;
\end{align} 

\begin{align}
    q(\rvx_{t-1} | \rvx_t, \rvx_0) &= \frac{q(\rvx_t | \rvx_{t-1}) q(\rvx_{t-1}| \rvx_0)}{q(\rvx_t | \rvx_0)} = \frac{\text{Cat}(\rvx_t; Q_{t}^\top\rvx_{t-1})\text{Cat}(\rvx_{t-1} ; \overline{Q}_{t-1}^\top\rvx_0)}{\text{Cat}(\rvx_t ; \overline{Q}_t^\top\rvx_0)}
    \nonumber \\
    &= \frac{\rvx_{t-1}^\top Q_{t} \rvx_{t} \cdot \rvx_{t-1}^\top\overline{Q}_{t-1}^\top\rvx_0}{\rvx_t^\top\overline{Q}_t^\top\rvx_0 } 
    = \rvx_{t-1}^\top \frac{Q_t \rvx_{t} \odot \overline{Q}_{t-1}^\top\rvx_0}{\rvx_t^\top\overline{Q}_t^\top\rvx_0 } = \text{Cat}(\rvx_{t-1};\frac{Q_t \rvx_{t} \odot \overline{Q}_{t-1}^\top\rvx_0}{\rvx_t^\top\overline{Q}_t^\top\rvx_0 })
\end{align}

\subsection{Simplification of Transition Matrix}
\label{ssec:simple_trans_matrix}

For \textit{any} transition matrices $Q_1,...,Q_T$, which however should be chosen such that every row of $\overline{Q}_t= \overline{Q}_{t|0}$ converge to the same known stationary distribution when $t$ becomes large (i.e. at $T$), let the known stationary distribution be $\rvm_0 \sim \text{Cat}(\rvm_0;\vm)$. Then, the constraint can be stated as 
{
\setlength{\abovedisplayskip}{4pt}
\setlength{\belowdisplayskip}{-4pt}
\begin{align}
    \lim_{t\rightarrow T} \overline{Q}_t  = \1 \vm^\top \;. \label{eq:stationary}
\end{align}
}

To achieve the desired convergence on nominal data (which is typical data type in edges and nodes of graphs), while keeping the flexibility of choosing any categorical stationary distribution $\rvm_0 \sim \text{Cat}(\rvm_0;\vm)$, we define $Q_t$ as 
{
\setlength{\abovedisplayskip}{4pt}
\setlength{\belowdisplayskip}{-4pt}
\begin{align}
    Q_t = \alpha_t I + (1-\alpha_t)\1\vm^\top \;, \label{eq:Qt}
\end{align}
}

where $\alpha_t \in [0, 1]$. 
This results in the accumulated transition matrix $\overline{Q}_{t|s} $ being equal to 
{
\setlength{\abovedisplayskip}{3pt}
\setlength{\belowdisplayskip}{-3pt}
\begin{align}
    \overline{Q}_{t|s} = \overline{\alpha}_{t|s} I + (1-\overline{\alpha}_{t|s} )\1 \vm^\top  \; 
   \forall t > s \;,\label{eq:Qbar_t_s}
\end{align}
}

where $\overline{\alpha}_{t|s} = \prod_{i=s+1}^t \alpha_i $. Note that $ \overline{\alpha}_{t} = \overline{\alpha}_{t|0} =\overline{\alpha}_{t|s} \overline{\alpha}_{s}$. We achieve 
%the constraint in 
\eqref{eq:stationary} by picking $\alpha_t$ such that $\lim_{t\rightarrow T} \overline{\alpha}_t  = 0$.

\subsection{Proof of Theorem \ref{thm:structural-equivalence}}
\label{apdx:proof-structural-equivalence}

We prove this for node case. For structurally equivalent edges, the analysis is the same. 
Assume node $i$ and node $j$ are structually equivalent, then we can find an automorphism $\sigma \in \text{Aut}(G)$ such that 
$\sigma(i) = j$. For any permutation $\mP \in \sP_{|G|}$ and an equivariant network $f$, we have 
\begin{align}
    f(\mP \star \ermG) = \mP \star f(\ermG)
\end{align}

Replace $\mP$ with $\sigma$, and using the fact that $\sigma \star \ermG = \ermG$. We can get 
\begin{align}
   f(\ermG) = f(\sigma \star \ermG) = \sigma \star f(\ermG)
\end{align}

Hence, we get $f(\ermG)_i = f(\ermG)_j$, that is two nodes $i$ and $j$ have the same representation.

\subsection{Proof of Exchangeable Probability}
\label{apdx:exchangeable}

To prove that $p_\theta(\ermG)$ is an exchangeable probability, we need to show that for any permutation operator $\mP$ with $\mP \star \ermG = (\mP \ermV, \mP\ermE \mP^\top)$, the probability satisfies $p_\theta(\mP \star\ermG) = p_\theta(\ermG)$. While $\mP$  is defined as a group acting on all nodes ($|V|$) in the graph $G$, this operator can also be applied directly to sub-components of G, in such a manner that it permutes only the elements within these components. We use $\mP$ to these sub-components directly in the following proof. 

Recall that \method define $p_\theta(\ermG)$ as 
\begin{align}
    p_\theta(\ermG) = \prod_{i=1}^{K_B} p_\theta\Big(\ermG[\gB_{1:i}] \setminus \ermG[\gB_{1:i-1}] \  \Big|\  \ermG[\gB_{1:i-1}] \Big).
\end{align}

Observe that for a graph $G$, $\gB_i$ represents a subset containing certain nodes from $G$. When $G$ is represented in its default order $\ermG$, the subset $\gB_i$ can be represented as a binary vector mask $\in\{0,1\}^{|\gV|}$, where a value of 1 at the $j$-th position indicates that the $j$-th node in $\ermG$ is included in the subset. Then $\ermG[\gB_i]$ can be view as indexing nodes and edges from $\ermG$ using mask $\gB_i$. As indexing operation is permutation equivariant, we have $\mP \star (\ermG[\gB_{1:i}]) = (\mP \star \ermG)[\mP \star \gB_{1:i}] $.
What is more,  $\gB_{i}$ is actually a function of $\ermG$ and can be represented as $\gB_{1:i}(\ermG)$. This function is actually permutation equivariant, such that $\gB_{i}(\mP \star \ermG) = \mP \star \gB_{i}(\ermG)$, as  $\gB_i (\ermG)$ is determined by algorithm 1 solely based on structural degree information (up to $K$ hops) of each node, where the structural feature is permutation equivariant. 

Then we have
\begin{align}
    &p_\theta(\mP \star \ermG) \nonumber\\
    & = \prod_{i=1}^{K_B} p_\theta\Big((\mP \star \ermG)[\gB_{1:i}(\mP \star \ermG)] \setminus (\mP \star \ermG)[\gB_{1:i-1}(\mP \star \ermG)] \  \Big|\  (\mP \star \ermG)[\gB_{1:i-1}(\mP \star \ermG)] \Big) \nonumber\\
    & = \prod_{i=1}^{K_B} p_\theta\Big((\mP \star \ermG)[\mP \star \gB_{1:i}] \setminus (\mP \star \ermG)[\mP \star \gB_{1:i-1}] \  \Big|\  (\mP \star \ermG)[\mP \star \gB_{1:i-1}] \Big)   \quad \text{(Prop. \ref{eq:order-equivariant})} \nonumber\\
    % &\omit\hfill foo\\ 
    & = \prod_{i=1}^{K_B} p_\theta\Big(\mP \star (\ermG[\gB_{1:i}]) \setminus \mP \star\ermG[\gB_{1:i-1}] \  \Big|\  \mP \star(\ermG[\gB_{1:i-1}]) \Big)  \quad \text{(Indexing's equivariance)} \nonumber\\
    & = \prod_{i=1}^{K_B} p_\theta\Big(\mP \star (\ermG[\gB_{1:i}] \setminus\ermG[\gB_{1:i-1}]) \  \Big|\  \mP \star(\ermG[\gB_{1:i-1}]) \Big)   \nonumber\\
    & = \prod_{i=1}^{K_B} p_\theta\Big(|\gB_i|\ \Big| \  \mP \star\ermG[\gB_{1:i-1}]  \Big)  
      p_\theta\Big(\mP \star (\ermG[\gB_{1:i}] \setminus\ermG[\gB_{1:i-1}]) \Big|\  \mP \star (\ermG[\gB_{1:i-1}]  \cup \emptyset[\gB_{1:i}]) \Big) \label{eq:equi-proof-0}
\end{align}

Notice that we are using a permutation invariant function to approximate $ p_\theta\Big(|\gB_i|\ \Big| \  \mP \star\ermG[\gB_{1:i-1}]  \Big)$, hence we have 
\begin{align}
     p_\theta\Big(|\gB_i|\ \Big| \  \mP \star\ermG[\gB_{1:i-1}]  \Big) = 
      p_\theta\Big(|\gB_i|\ \Big| \  \ermG[\gB_{1:i-1}]  \Big). \label{eq:equi_proof-n1}
\end{align}
For the second part, notice that 
\begin{align}
     &p_\theta\Big(\ermG[\gB_{1:i}] \setminus\ermG[\gB_{1:i-1}] \Big|\  \ermG[\gB_{1:i-1}]  \cup \emptyset[\gB_{1:i}] \Big)
     = \nonumber\\
     &\int p_\theta\Big( \ermH[\gB_{1:i-1}]  \cup \big(\ermG[\gB_{1:i}] \setminus\ermG[\gB_{1:i-1}] \big)  \Big|\  \ermG[\gB_{1:i-1}]  \cup \emptyset[\gB_{1:i}] \Big) d \ermH[\gB_{1:i-1}]
\end{align}
Then we have 
\begin{align}
     &p_\theta\Big( \mP \star (\ermG[\gB_{1:i}] \setminus\ermG[\gB_{1:i-1}]) \Big|\  \mP \star ( \ermG[\gB_{1:i-1}]  \cup \emptyset[\gB_{1:i}] ) \Big) \nonumber\\
     &= 
     \int p_\theta\Big( \ermH[\gB_{1:i-1}]  \cup \mP \star \big(\ermG[\gB_{1:i}] \setminus\ermG[\gB_{1:i-1}] \big)  \Big|\ 
     \mP \star (\ermG[\gB_{1:i-1}]  \cup \emptyset[\gB_{1:i}] ) \Big) d \ermH[\gB_{1:i-1}] \nonumber\\
    &= 
     \int p_\theta\Big( \mP \star\ermH[\gB_{1:i-1}]  \cup \mP \star \big(\ermG[\gB_{1:i}] \setminus\ermG[\gB_{1:i-1}] \big)  \Big|\ 
     \mP \star (\ermG[\gB_{1:i-1}]  \cup \emptyset[\gB_{1:i}] ) \Big) d \mP \star\ermH[\gB_{1:i-1}] \nonumber\\   
     &= \int p_\theta\Big( \mP \star \big ( \ermH[\gB_{1:i-1}]  \cup  \big(\ermG[\gB_{1:i}] \setminus\ermG[\gB_{1:i-1}] \big) \big)  \Big|\ 
     \mP \star (\ermG[\gB_{1:i-1}]  \cup \emptyset[\gB_{1:i}] ) \Big) d \ermH[\gB_{1:i-1}] \label{eq:equi-proof-1}
\end{align}

Now notice that the probability $p_\theta\Big( \mP \star \big ( \ermH[\gB_{1:i-1}]  \cup  \big(\ermG[\gB_{1:i}] \setminus\ermG[\gB_{1:i-1}] \big) \big)  \Big|\ 
     \mP \star (\ermG[\gB_{1:i-1}]  \cup \emptyset[\gB_{1:i}] ) \Big)$ is actually a conditional distribution from one graph to another graph (with same size), and we can simplify it as $p_\theta(\ermH | \ermG)$. As this part is modeled by diffusion model, we prove that this function is permutation equivariant. That is, for any permutation $\mP$, 
\begin{align}
    p_\theta(\ermH | \ermG) = p_\theta(\mP \star \ermH |\mP \star \ermG) \label{eq:equi-proof-2}
\end{align}
\begin{proof}
 Our diffusion process is using an equivariant model through a markov chain that from $ \ermG \text{ to }  \ermH_T$, and then from $ \ermH_T \text{ to } \ermH_{T-1}$, ..., $ \ermH_{t} \text{ to }  \ermH_{t-1}$, until $ \ermH_{1} \text{ to }  \ermH_{0} = \ermH$. Then we have 
\begin{align}
 p_\theta(\ermH | \ermG) =  p_\theta(\ermH_0 | \ermG) = \int p_\theta(\ermH_T | \ermG) \prod_{t=1}^T p_\theta(\ermH_t | \ermH_{t-1}) d \ermH_{1:T}
\end{align}

Because we are using an shared permutation equivariant model to model all  $p_\theta(\ermH_t | \ermH_{t-1})$, we have $p_\theta(\ermH_t | \ermH_{t-1}) = p_\theta(\mP \star \ermH_t | \mP \star \ermH_{t-1}) \forall t, \mP$. Also, because we have chosen the distribution $p_\theta(\ermH_T | \ermG)$, such that all conditioned part from $\gB_{1:i}$ are the same with input G, and all other left elements are sampled from isotrophic noise, the distribution  $p_\theta(\ermH_T | \ermG)$ is also equivariant with  $p_\theta(\ermH_T | \ermG) =  p_\theta( \mP \star \ermH_T |\mP \star \ermG). $ 
Take these properties back we have 
\begin{align}
     p_\theta(\mP \star \ermH | \mP \star \ermG) &= \int p_\theta(\ermH_T | \mP \star \ermG) p_\theta(\ermH_1 | \mP \star \ermH_{0}) \prod_{t=2}^T p_\theta(\ermH_t | \ermH_{t-1}) d \ermH_{1:T} \nonumber\\
     % & = \int p_\theta(\mP \star \ermH_T | \mP \star \ermG) p_\theta( \mP \star \ermH_1 | \mP \star \ermH_{0}) \prod_{t=2}^T p_\theta(\mP \star \ermH_t | \mP \star \ermH_{t-1}) d \mP \star \ermH_{1:T} \\
     & = \int p_\theta(\mP \star \ermH_T | \mP \star \ermG) p_\theta( \mP \star \ermH_1 | \mP \star \ermH_{0}) \prod_{t=2}^T p_\theta(\mP \star \ermH_t | \mP \star \ermH_{t-1}) d\ermH_{1:T} \nonumber\\
     & = \int p_\theta(\ermH_T | \ermG) \prod_{t=1}^T p_\theta(\ermH_t | \ermH_{t-1}) d \ermH_{1:T} \nonumber\\
     & =  p_\theta(\ermH | \ermG)
\end{align}
\end{proof}

With the conclusion from \eqref{eq:equi-proof-2}, we can apply it to \eqref{eq:equi-proof-1} 
\begin{align}
    &p_\theta\Big( \mP \star (\ermG[\gB_{1:i}] \setminus\ermG[\gB_{1:i-1}]) \Big|\  \mP \star ( \ermG[\gB_{1:i-1}]  \cup \emptyset[\gB_{1:i}] ) \Big) \nonumber\\
    &= \int p_\theta\Big( \mP \star \big ( \ermH[\gB_{1:i-1}]  \cup  \big(\ermG[\gB_{1:i}] \setminus\ermG[\gB_{1:i-1}] \big) \big)  \Big|\ 
     \mP \star (\ermG[\gB_{1:i-1}]  \cup \emptyset[\gB_{1:i}] ) \Big) d \ermH[\gB_{1:i-1}] \nonumber\\
     & =  \int p_\theta\Big(   \ermH[\gB_{1:i-1}]  \cup  \big(\ermG[\gB_{1:i}] \setminus\ermG[\gB_{1:i-1}] \big)  \Big|\ 
     \ermG[\gB_{1:i-1}]  \cup \emptyset[\gB_{1:i}] \Big) d \ermH[\gB_{1:i-1}] \nonumber\\
     & = p_\theta\Big(\ermG[\gB_{1:i}] \setminus\ermG[\gB_{1:i-1}] \Big|\  \ermG[\gB_{1:i-1}]  \cup \emptyset[\gB_{1:i}] \Big) \label{eq:equi_proof-n2}
\end{align}

Apply \eqref{eq:equi_proof-n1} and \eqref{eq:equi_proof-n2} to \eqref{eq:equi-proof-0}, we can finalize our proof that 
\begin{align}
     p_\theta(\mP \star \ermG)& = \prod_{i=1}^{K_B} p_\theta\Big(|\gB_i|\ \Big| \  \mP \star\ermG[\gB_{1:i-1}]  \Big)  
      p_\theta\Big(\mP \star (\ermG[\gB_{1:i}] \setminus\ermG[\gB_{1:i-1}]) \Big|\  \mP \star (\ermG[\gB_{1:i-1}]  \cup \emptyset[\gB_{1:i}]) \Big) \nonumber\\
      & = \prod_{i=1}^{K_B} p_\theta\Big(|\gB_i|\ \Big| \  \ermG[\gB_{1:i-1}]  \Big)  
      p_\theta\Big(\ermG[\gB_{1:i}] \setminus\ermG[\gB_{1:i-1}] \Big|\  \ermG[\gB_{1:i-1}]  \cup \emptyset[\gB_{1:i}] \Big) \nonumber\\
      & =  \prod_{i=1}^{K_B} p_\theta\Big(\ermG[\gB_{1:i}] \setminus \ermG[\gB_{1:i-1}] \  \Big|\  \ermG[\gB_{1:i-1}] \Big) \nonumber = p_\theta(\ermG) 
\end{align}

Hence, our method \method is a permutation-invariant probability model.

\subsection{Definition of Graph Transformation}
\label{apdx:definition}

\begin{definition}[Graph Transformation]
Given a labeled graph \(\ermG = (\ermV_G, \ermE_G)\) and a labeled target graph \(\ermH = (\ermV_H, \ermE_H)\) that have the same number of nodes, the graph transformation function \(f: \ermG \rightarrow \ermH\) is defined such that f(\ermG) = \ermH. 
\end{definition}

Notice that while the graph transformation function requires input and output graph have the same graph size, this function is general enough to accommodate any changing of size operation. Specifically, if $\ermG$ has more nodes than $\ermH$, the function must assign an 'empty' token to nodes and edges within $\ermG$ that do not correspond to those in $\ermH$. Conversely, if $\ermG$ has fewer nodes than $\ermH$, the function will augment $\ermG$ with 'empty' nodes and edges until it matches the size of $\ermH$. Subsequently, it transforms this augmented version of $\ermG$ to match $\ermH$.

\begin{definition}[Equivariant Graph Transformation]

An equivariant graph transformation $f$ is a graph transformation that satisfies $f(\mP \star \ermG) = f(\mP \star \ermH)$ for any permutation operation $\mP$.
\end{definition}

\subsection{Algorithms of Training and Generation}
\label{apdx:train-inference}

We present blocksize prediction algorithm in Algo. \ref{alg:train-blocksize} and denoising diffusion algorithm  in Algo. \ref{alg:train-diffusion}.
Notice while the blocksize network $g_\theta$ in Algo. \ref{alg:train-blocksize} and denoising diffusion network $f_\theta$ in Algo. \ref{alg:train-diffusion} use the same subscript parameter $\theta$, the $\theta$ essentially represents concatenation of $g$ and $f$'s parameters. 

\input{Algorithms/blcoksize-train}
\input{Algorithms/diffusion-train}
\input{Algorithms/generation}

\subsection{Visualization of the PPGN-Transformer Block}
\label{apdx:ppgn-transformer}

\begin{figure}[h]
\vspace{-0.1in}
    \centering    \includegraphics[width=0.85\linewidth]{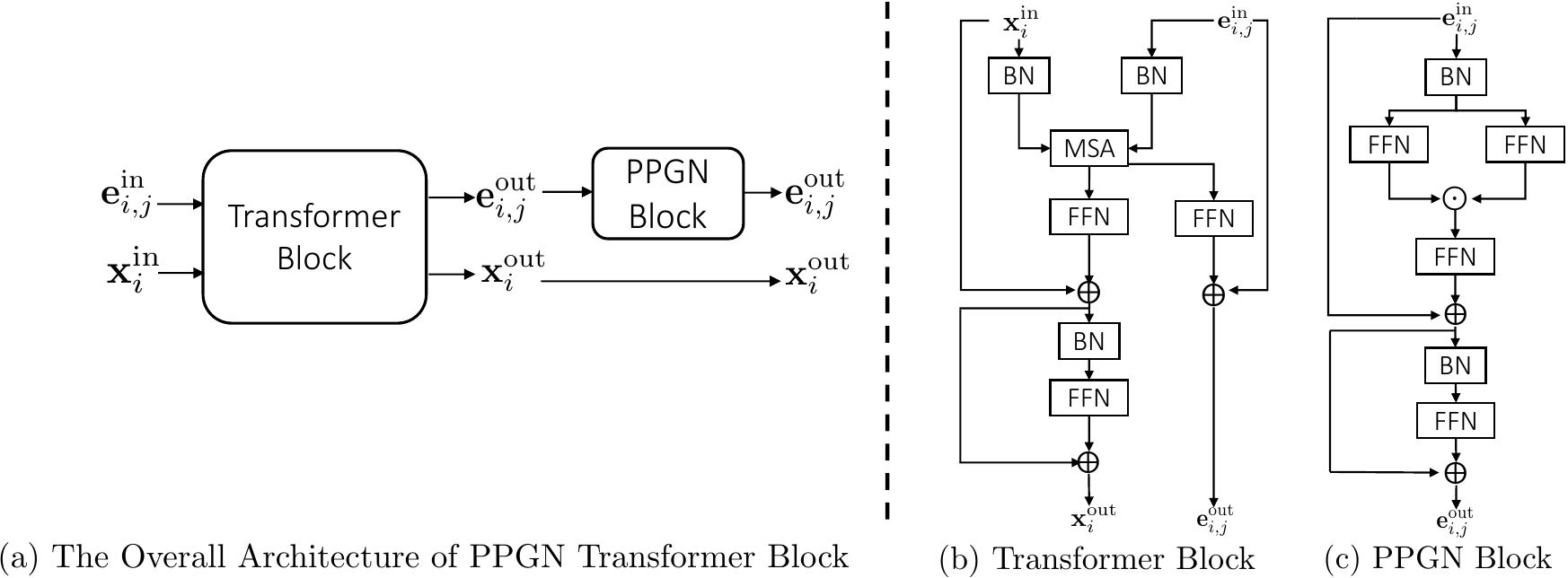}
    \caption{The Architecture of the PPGN-Transformer Block. In (b) and (c) we provide illustrations of how edge and node features are processed through Transformer and PPGN blocks.}
    \label{fig:ppgn_arch}
    %\vspace{-0.1in}
\end{figure}

\subsection{Details and Derivations of Causal Matrix-Matrix Product}
\label{apdx:causal}

\subsubsection{Derivation of causal version matrix product}

For any matrix $\mX$, let $\mX_i$ denotes the $i$-th row of $\mX$, and $\mX_{:,i}$ denotes $i$-th column of $\mX$. Let $\langle \cdot,\cdot \rangle$ denotes vector inner product. 
For normal matrix product $\mA\mB$, we know that 
\begin{align}
 (\mA\mB)_{ij} =  \langle \mA_i,\mB_{:,j} \rangle \label{eq:ab}
\end{align}
In our causal setting with block ID introduced in main context, the $(i,j)$ position will have its block ID being $\max( \text{block\_ID}(i), \text{block\_ID}(j) )$. Hence, the product in \eqref{eq:ab} should not use any information outside of block whose ID is larger than  $\max( \text{block\_ID}(i), \text{block\_ID}(j) )$. Let $\mO_{ij}$ be the needed safe output at position $(i,j)$ of the matrix-matrix product, one can verify that the following uses all information within useable blocks without any information leakage. 
\begin{align}
   \mO_{ij} &= \langle \mA_i \odot (\mM_i \text{ or } \mM_{j})  , \mB_{:,j}  \rangle  \quad \text{(where or denotes elementwise or operation)} \nonumber \\
   &= \langle \mA_i \odot (\mM_i +   \mM_{j} - \mM_i \odot \mM_{j} ) , \mB_{:,j}  \rangle  \nonumber \\ 
   &= \langle \mA_i \odot \mM_i,  \mB_{j}  \rangle + 
   \langle \mA_i,  \mB_{:,j} \odot \mM_j  \rangle  - 
   \langle \mA_i \odot \mA_i,  \mB_{:,j} \odot \mM_j \rangle  \nonumber\\
   & = \langle \mA_i \odot \mM_i,  \mB_{j}  \rangle + 
   \langle \mA_i,  \mB_{:,j} \odot \mM^\top_{:,j}  \rangle  - 
   \langle \mA_i \odot \mA_i,  \mB_{:,j} \odot \mM^\top_{:,j}  \rangle \label{eq:causal_ab}
\end{align}

Where mask matrix $\mM$ satisfies $\mM_{i,j} = 1 $ if $\text{block\_ID}(\text{node } i) \ge \text{block\_ID}(\text{node }j) $ else 0.

Based on \eqref{eq:causal_ab}, we can rewrite it to matrix operation such that 
\begin{align}
\mO = (\mA \odot \mM)\mB + \mA(\mB \odot \mM^\top) - (\mA \odot \mM)(\mB \odot \mM^\top)
\end{align}
Where the matrix $\mO$'s $(i,j)$ position value is just $\mO_{ij}$ in \eqref{eq:causal_ab}. Hence we have derived the equation in \eqref{eq:causal_matrix_prod}.

\subsubsection{Discussion of expressivity}
While the causal matrix-matrix product makes the training more efficient with parallel support, we have to acknowledge that it is highly possible that its expressiveness in graph distinguishing ability is reduced. Hence the causal PPGN should have less expressiveness than the normal PPGN. In fact, we have found that for symmetry-rich datasets like \grid, sequential training of \method's loss is easier to minimize than parallel training of \method. 

\subsubsection{Additional details of parallel training}

The previous discussion mainly focuses on preventing information leakage, which is the most critical aspect of causal parallel training. Another essential component of parallel training is the ability to output all predictions for subsequent blocks simultaneously. For GPT, no modifications are necessary for predicting all next tokens, as the next token can simply use the position from the previous token. However, when predicting the next block given all previous blocks, it's not feasible to use any positions within previous blocks to hold the prediction for the next block due to differences in size and lack of alignment. Therefore, as shown in \eqref{eq:p_conditional_decom}, a virtual block needs to be augmented to serve as a prediction placeholder for the next block. For parallel prediction, a virtual block will be augmented for each block in the graph, with each virtual block having the same block ID as the corresponding original block. To summarize, for a graph with $N$ nodes, it needs to be expanded to $2N$ nodes by adding $N$ virtual nodes for parallel training.

\subsection{Experiment Details}
\label{ssec:exp_details}
\input{Tables/data}

We use a single RTX-A6000 GPU for all experiments. For discrete diffusion, we follow \cite{zhao2024improving}'s basic discrete diffusion implementation, where exponential moving average is not enabled. For model implementation, we use Pytorch Geometric \cite{pyg}, and we implement our combination of PPGN and Transformer by referencing the code in \citet{PPGN} and \citet{ma2023GraphInductiveBiases}. Additionally, we use Pytorch Lightning \cite{falcon2019pytorch} for training and keeping the code clean. We use Adam optimizer with cosine decay learning rate scheduler to train. 
For diffusion and blocksize prediction, we also input the embedding of block id and node degree as additional feature, as we have block id computed in preprocessing. Additionally, it is important to observe that nodes within the same block all have the same degree. At diffusion stage, we conditional on the predicted degree for the next block to train diffusion model, where the degree is predicted along with the block size using the same network.
We provide source code and all configuration files at \url{https://github.com/LingxiaoShawn/Pard}.

\subsection{Ablations and Runtime Measure}
\label{apdx:ablation}

\textbf{How does the sampling steps for diffusion models in \method impact the performance of graph generation?} To answer it, we conducted a detailed ablation on number of diffusion steps in QM9, with results in Table. \ref{tab:ablation_qm9}.  To summarize, too small a number of steps hurt the performance, while too large doesn’t help improve the performance further. See the table below. Within the table, we have emphasized a configuration that employs just 140 total diffusion steps. This amount is considerably less than the steps used by DiGress, yet it dramatically outperforms DiGress.
\input{Tables/ablation}

\textbf{While graphs are trained with a single, fixed generation path based on $\phi$, is the same path being used in generation?}  We acknowledge that although each graph undergoes a single decomposition block sequence during training, it is possible for the generation path to differ from the training path. This discrepancy is attributed to a phenomenon known as \textit{exposure bias}, a challenge encountered in autoregressive methods and imitation learning alike. Despite this, we have conducted empirical analysis to estimate the likelihood of divergence between the generation and training paths. By sampling 2,000 graphs generated by the model trained on the QM9 dataset, we tracked the generation path for each graph and compared it to the training path defined by the decomposition block sequence algorithm. Our findings indicate that \textbf{94.7\%} of the generated samples followed the exact same path as their corresponding training path.

\textbf{Runtime comparison.} To show that our method doesn’t introduce much runtime overhead, we report the training time in Table \ref{tab:training_time} used to achieve the number reported on two datasets: QW9 and Moses. The training time for DiGress is mentioned in their github. Notice that the training time is also affected by the machine, we use GPU A6000 in all experiments. 

\begin{table}[ht]
\centering
\caption{Training time comparison on QM9 and Moses datasets}
\label{tab:training_time}
\begin{tabular}{l|c|c}
\toprule
Training Time & QM9 & Moses \\
\midrule
DiGress & $\sim$6h & $\sim$7 days \\
\method & $\sim$12h & $\sim$4 days \\
\bottomrule
\end{tabular}
\end{table}

\subsection{Visualization of Generation}

\subsubsection{QM9}
\begin{figure}[H]
    \centering
    \includegraphics[width=0.7\textwidth]{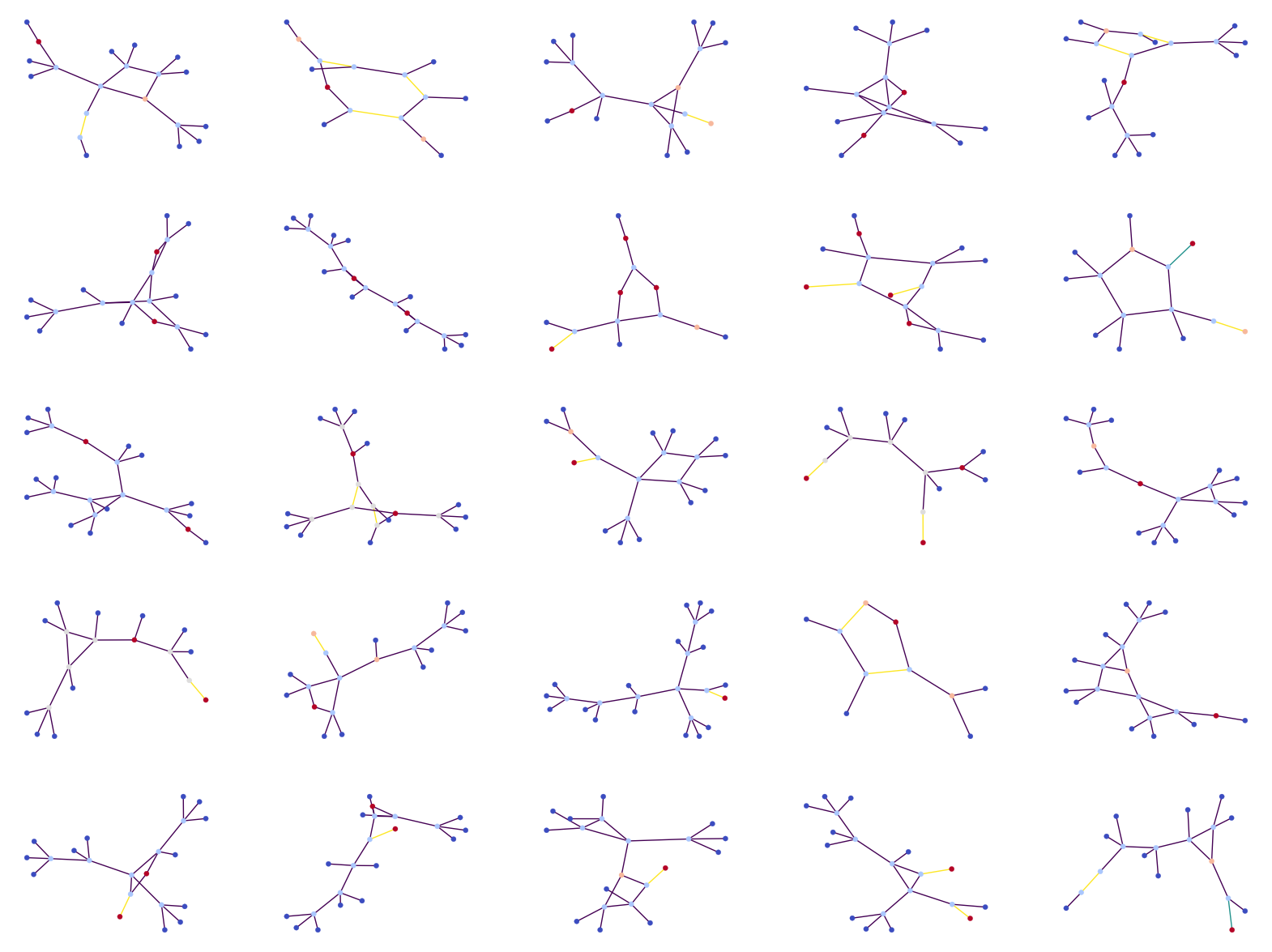}
    \caption{Non curated QM9 (with explicit hydrogens) graphs generated  from the \method trained with 20 steps per block.}
    \label{fig:enter-label}
\end{figure}

\subsubsection{Grid}
\begin{figure}[H]
    \centering
    \includegraphics[width=0.7\textwidth]{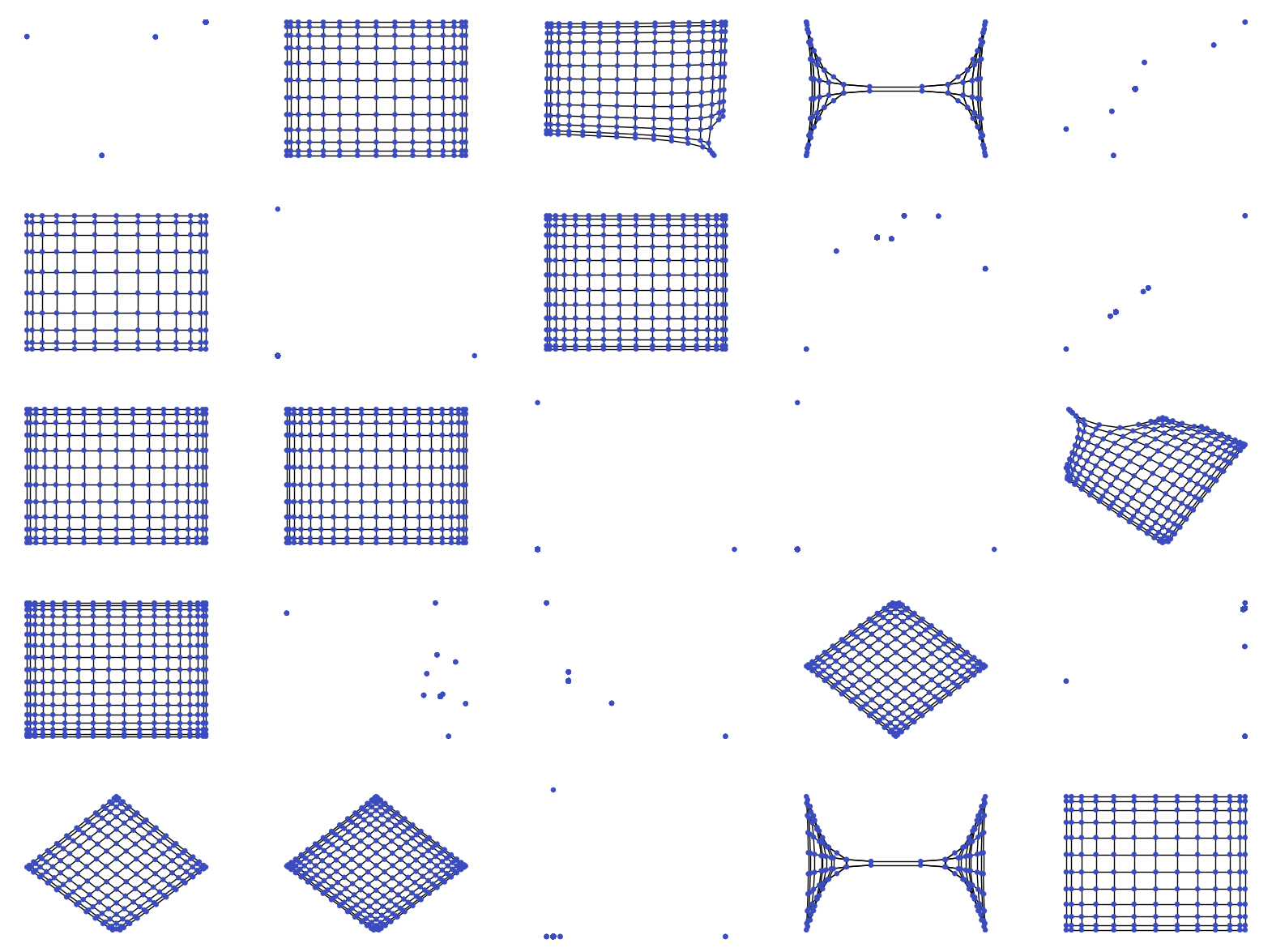}
    \caption{Non curated grid graphs generated  from the \method trained with 50 steps per block.}
    \label{fig:enter-label}
\end{figure}

\begin{figure}[H]
    \centering
    \includegraphics[width=0.7\textwidth]{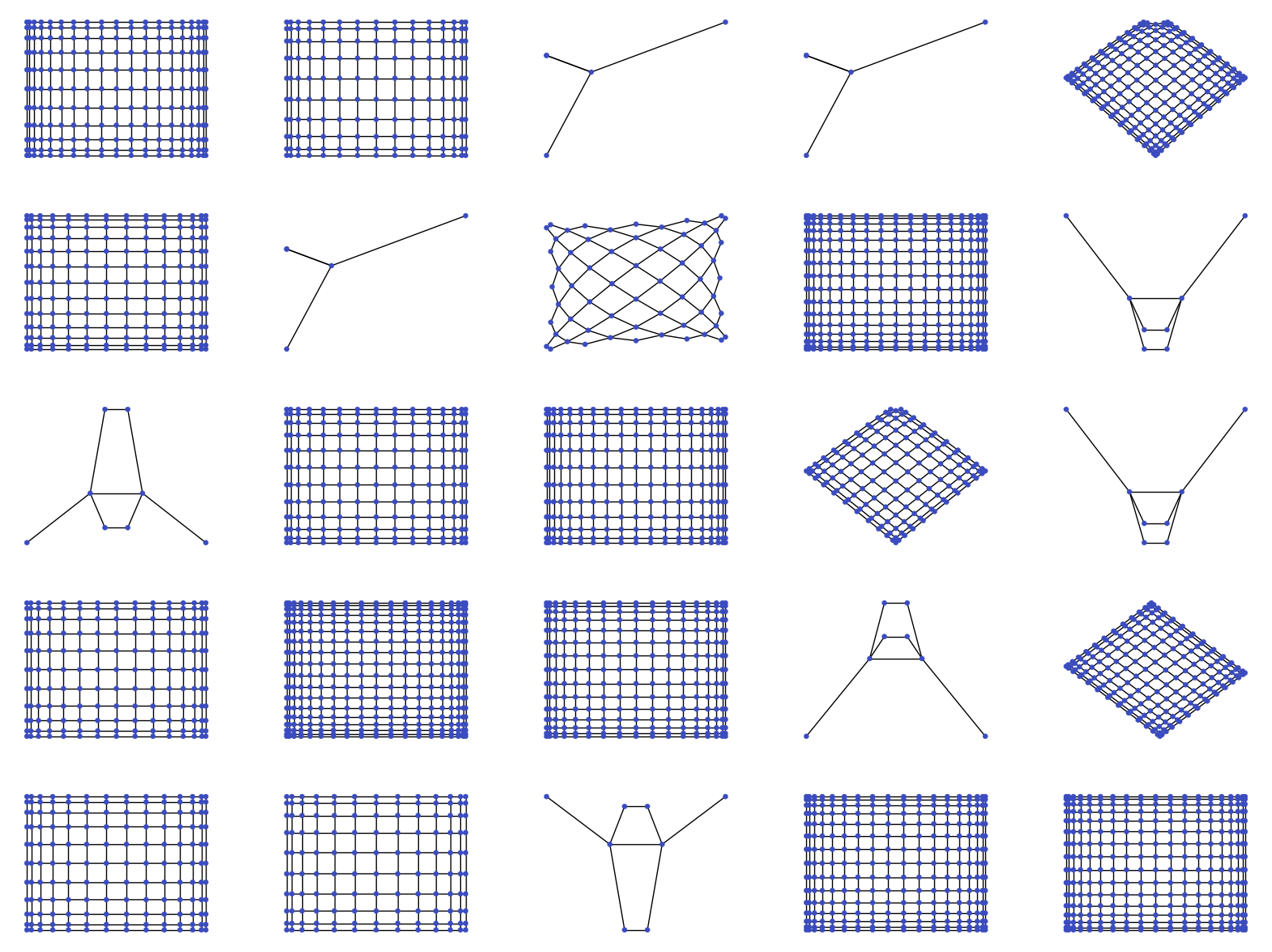}
    \caption{Non curated grid graphs generated  from the \method (with eigenvector) trained with 50 steps per block. }
    \label{fig:enter-label}
\end{figure}

\begin{table}[]
    \centering
    \begin{tabular}{c|c|c|c}
    \toprule
   \grid & \textbf{Deg.} & \textbf{Clus.} & \textbf{Orbit} \\
   \midrule
     EDP-GNN    & 0.455 & 0.238 & 0.328\\
     GDSS & 0.111 & 0.005 & 0.070 \\
     \method    & 0.028 & 0.002 & 0.029 \\
     \method with EigenVec.     & 0.053 & 0 & 0.008 \\
     \bottomrule
    \end{tabular}
    \caption{Performance comparison of diffusion methods on \grid.}
    \label{tab:grid-table}
\end{table}

%% file: Algorithms/blcoksize-train.tex
\begin{algorithm}[H]
\caption{Train blocksize distribution $p_\theta\big(|\gB_i| \ \big| \  \ermG[\gB_{1:i-1}] \big)$}
\label{alg:train-blocksize}
\begin{algorithmic}[1]
\STATE \textbf{Input:} $\ermG$, maximum hop $K_h$, a network $g_\theta$ that takes a graph as input and output graph wise prediction.
\STATE Get structural partial order function $\phi$ of $\ermG$ from Algo.\ref{alg:partial_order}.
\STATE Using $\phi$ to get the sequence of node blocks $[\gB_1,...,\gB_{K_B}]$ for $\ermG$.
\STATE Minimize $\sum_{i=1}^{K_B} \text{CrossEntropy}(g_\theta(\ermG[\gB_{i}]),|\gB_{i+1}|) $, with $|\gB_{K_B+1}|=0$
\end{algorithmic}
\end{algorithm}

%% file: Algorithms/diffusion-train.tex
\begin{algorithm}[H]
\caption{Train denoising diffusion for distribution $p_\theta\Big(\ermG[\gB_{1:i}] \setminus \ermG[\gB_{1:i-1}] \  \Big|\  \ermG[\gB_{1:i-1}]  \cup \emptyset[\gB_{1:i}] \Big)$}
\label{alg:train-diffusion}
\begin{algorithmic}[1]
\STATE \textbf{Input:} $\ermG$, max time T, maximum hop $K_h$, a network $f_\theta$ that inputs a graph and outputs nodes and edges predictions.
\STATE Get structural partial order function $\phi$ of $\ermG$ from Algo.\ref{alg:partial_order}.
\STATE Using $\phi$ to get the sequence of node blocks $[\gB_1,...,\gB_{K_B}]$ for $\ermG$.
\STATE Sample $t \sim U(1,...,T)$
\FOR{$i=1,..., K_B$}  
\STATE  $M \leftarrow $ indice mask of $\ermG[\gB_{1:i}] \setminus \ermG[\gB_{1:i-1}]$
\STATE Sample a noise graph $\tilde{\ermG}[\gB_{1:i}]$ from $q_{t|0}(\ermG[\gB_{1:i}])$ according to \eqref{eq:t|0}
\STATE $\tilde{\ermG}[\gB_{1:i}] \leftarrow M \odot \tilde{\ermG}[\gB_{1:i}] + (1-M) \odot \ermG[\gB_{1:i}]  $
\STATE $X \leftarrow f_\theta(\tilde{\ermG}[\gB_{1:i}] ) \odot M $
\STATE $Y \leftarrow \ermG[\gB_{1:i}] \odot M$
\STATE $l_i \leftarrow \Ls_t (X,Y) + 0.1 *\Ls_t^{CE}(X,Y)$, using 
$\Ls_t$ in \eqref{eq:vlb} and  $\Ls_t^{CE}$ in \eqref{eq:ce}.
\ENDFOR
\STATE Minimize $\sum_{i=1}^{K_B} l_i$.   \qquad(The for loop can be parallelized. )
\end{algorithmic}
\end{algorithm}

%% file: Algorithms/generation.tex
\begin{algorithm}[H]
\caption{Generation}
\label{alg:generation}
\begin{algorithmic}[1]
\STATE \textbf{Input:} blocksize model $g_\theta$, diffusion model $f_\theta$; first blocksize distribution from TrainSet. 
\STATE $\ermG \leftarrow \emptyset$; $i \leftarrow 1$
\STATE Sample $n$ from the first block's size distribution. 
\WHILE{$n > 0$}
\STATE Add a new block $\gB_i$ with $n$ nodes into $\ermG$ 
\STATE $M \leftarrow $ indice mask of $\ermG[\gB_{1:i}] \setminus \ermG[\gB_{1:i-1}]$
\STATE $\tilde{\ermG} \leftarrow$ For nodes and edges within $M$, sample from noise $\vm_n$ and $\vm_e$.
\FOR{$j=1:T$}
    \STATE \text{p} $\leftarrow f_\theta(\tilde{\ermG})$
    \STATE $S \leftarrow$ Sample according to \text{p}
    \STATE $\tilde{\ermG} \leftarrow M \odot S + (1-M)\odot \tilde{\ermG}$
\ENDFOR
\STATE $\ermG \leftarrow \tilde{\ermG}$
\STATE $n \leftarrow$ Sample from $g_\theta(\ermG)$
\STATE $i \leftarrow i + 1$
\ENDWHILE
\STATE \textbf{Return:}  $\ermG$
\end{algorithmic}
\end{algorithm}

%% file: Tables/data.tex
\begin{table}[H]
\setlength{\tabcolsep}{5pt}
\centering
\caption{Dataset summary}
\begin{tabular}{llll}
\toprule
\textbf{Name} & \textbf{\#Graphs} &$|V|_{\text{avg}}$ & $|E|_{\text{avg}}$ \\
\midrule
  \qm   &   133,885   &         9        &             19       \\
  \zinc   &    249,455      &    23        &     50         \\
  \moses   &    1,936,963 &     22      &       47             \\
  \hline
 {\sc Community-S}    &   200   &       15.8     &     75.5          \\
\cave                 &   200  &        13.9    &        68.8        \\
\cora                 &   18,850   &       51.9     &   121.2            \\
 \breast              &    100  &      55.7      &        117.0       \\
 \grid & 100 & 210 & 783.0\\
 \bottomrule
\end{tabular}
\label{tab:data}
\end{table}

%% file: Tables/ablation.tex
\begin{table}[ht]
\centering
\caption{Comparison of different models on QM9 dataset}
\label{tab:ablation_qm9}
\begin{tabular}{l|c|c|c|c|c|c}
\toprule
Ablation on QM9 & \multicolumn{5}{|c|} \method  & DiGress \\
\midrule
Average number of blocks & 7 & 7 & 7 & 7 & 7 & 1 \\
Number of Diffusion Steps Per Block & 10 & \textbf{20} & 40 & 70 & 100 & 500 \\
Total Number of Diffusion Steps & 70 & \textbf{140} & 280 & 490 & 700 & 500 \\
Validity & 0.9419 & 0.9708 & 0.971 & 0.973 & 0.9755 & 95.4 \\
Uniqueness & 0.9676 & 0.9605 & 0.9579 & 0.9544 & 0.9609 & 97.6 \\
Mol stability & 0.8094 & 0.8627 & 0.8625 & 0.8589 & 0.8561 & 79.8 \\
Atom Stability & 0.9767 & 0.9847 & 0.9844 & 0.9839 & 0.9837 & 98.1 \\
\bottomrule
\end{tabular}
\end{table}

%% file: 9-checklist.tex
\begin{enumerate}

\item {\bf Claims}
    \item[] Question: Do the main claims made in the abstract and introduction accurately reflect the paper's contributions and scope?
    \item[] Answer: \answerYes{} % Replace by \answerYes{}, \answerNo{}, or \answerNA{}.
    \item[] Justification: \textcolor{blue}{We propose a new graph diffusion method that integrates autoregressive graph generation with diffusion. We showcase our method in the main paper with thorough experiments.}

\item {\bf Limitations}
    \item[] Question: Does the paper discuss the limitations of the work performed by the authors?
    \item[] Answer: \answerYes{}% Replace by \answerYes{}, \answerNo{}, or \answerNA{}.
    \item[] Justification: \textcolor{blue}{We briefly discuss the limitations of PARD with respect to long training and sampling times, but we have pointed out potential improvements with for parallel training in causal transformers.}
    

\item {\bf Theory Assumptions and Proofs}
    \item[] Question: For each theoretical result, does the paper provide the full set of assumptions and a complete (and correct) proof?
    \item[] Answer: \answerYes{} % Replace by \answerYes{}, \answerNo{}, or \answerNA{}.
    \item[] Justification: \textcolor{blue}{All our theorems, formulas and proofs in the paper are numbered and referenced. The assumptions are clearly stated. The proofs are shown in main paper and appendix.}

    \item {\bf Experimental Result Reproducibility}
    \item[] Question: Does the paper fully disclose all the information needed to reproduce the main experimental results of the paper to the extent that it affects the main claims and/or conclusions of the paper (regardless of whether the code and data are provided or not)?
    \item[] Answer: \answerYes{} % Replace by \answerYes{}, \answerNo{}, or \answerNA{}.
    \item[] Justification: \textcolor{blue}{We introduce a new algorithm and architecture for graph generation tasks. The detailed explanations on the algorithm, and the experiment details are clearly shown in the appendix.}

\item {\bf Open access to data and code}
    \item[] Question: Does the paper provide open access to the data and code, with sufficient instructions to faithfully reproduce the main experimental results, as described in supplemental material?
    \item[] Answer: \answerYes{} % Replace by \answerYes{}, \answerNo{}, or \answerNA{}.
    \item[] Justification: \textcolor{blue}{We have publicized our training repository with an anonymous link referenced in the paper.}

\item {\bf Experimental Setting/Details}
    \item[] Question: Does the paper specify all the training and test details (e.g., data splits, hyperparameters, how they were chosen, type of optimizer, etc.) necessary to understand the results?
    \item[] Answer: \answerYes{} % Replace by \answerYes{}, \answerNo{}, or \answerNA{}.
    \item[] Justification: \textcolor{blue}{We use the same test bed as our benchmarking methods, so the dataset split and evaluation codes are publicly available. The hyperparameter configurations are discussed in the appendix as well as in our code library. }

\item {\bf Experiment Statistical Significance}
    \item[] Question: Does the paper report error bars suitably and correctly defined or other appropriate information about the statistical significance of the experiments?
    \item[] Answer: \answerYes{} % Replace by \answerYes{}, \answerNo{}, or \answerNA{}.
    \item[] Justification: \textcolor{blue}{We provide experiment results and evaluations over 10,000-25,000 sampled graphs, so the numbers are of high confidence.}

\item {\bf Experiments Compute Resources}
    \item[] Question: For each experiment, does the paper provide sufficient information on the computer resources (type of compute workers, memory, time of execution) needed to reproduce the experiments?
    \item[] Answer: \answerYes{}
    \item[] Justification: \textcolor{blue}{We provide the memory, running time, sampling time discussions in the main paper and in appendix. All experiments are running on A6000 GPUs.}
    
\item {\bf Code Of Ethics}
    \item[] Question: Does the research conducted in the paper conform, in every respect, with the NeurIPS Code of Ethics \url{https://neurips.cc/public/EthicsGuidelines}?
    \item[] Answer: \answerYes{} % Replace by \answerYes{}, \answerNo{}, or \answerNA{}.
    \item[] Justification: \textcolor{blue}{The authors have reviewed the code of ethics.The paper and the code base preserve anonymity.}

\item {\bf Broader Impacts}
    \item[] Question: Does the paper discuss both potential positive societal impacts and negative societal impacts of the work performed?
    \item[] Answer: \answerNA{} 
    \item[] Justification: \textcolor{gray}{There is no societal impact of the work performed.}
    
\item {\bf Safeguards}
    \item[] Question: Does the paper describe safeguards that have been put in place for responsible release of data or models that have a high risk for misuse (e.g., pretrained language models, image generators, or scraped datasets)?
    \item[] Answer:\answerNA{} 
    \item[] Justification: \textcolor{gray}{The paper poses no such risks.}

\item {\bf Licenses for existing assets}
    \item[] Question: Are the creators or original owners of assets (e.g., code, data, models), used in the paper, properly credited and are the license and terms of use explicitly mentioned and properly respected?
    \item[] Answer:  \answerYes{}% Replace by \answerYes{}, \answerNo{}, or \answerNA{}.
    \item[] Justification: \textcolor{blue}{All our datasets and benchmarking methods are from publicly available libraries under proper licenses.}

\item {\bf New Assets}
    \item[] Question: Are new assets introduced in the paper well documented and is the documentation provided alongside the assets?
    \item[] Answer: \answerYes{} % Replace by \answerYes{}, \answerNo{}, or \answerNA{}.
    \item[] Justification: \textcolor{blue}{We provide anonymous code base. }

\item {\bf Crowdsourcing and Research with Human Subjects}
    \item[] Question: For crowdsourcing experiments and research with human subjects, does the paper include the full text of instructions given to participants and screenshots, if applicable, as well as details about compensation (if any)? 
    \item[] Answer: \answerNA{}
    \item[] Justification: \textcolor{gray}{The paper does not involve crowdsourcing nor research with human subjects.}

\item {\bf Institutional Review Board (IRB) Approvals or Equivalent for Research with Human Subjects}
    \item[] Question: Does the paper describe potential risks incurred by study participants, whether such risks were disclosed to the subjects, and whether Institutional Review Board (IRB) approvals (or an equivalent approval/review based on the requirements of your country or institution) were obtained?
    \item[] Answer: \answerNA{} % Replace by \answerYes{}, \answerNo{}, or \answerNA{}.
    \item[] Justification: \textcolor{gray}{The paper does not involve crowdsourcing nor research with human subjects.}

\end{enumerate}